%% file: main.tex
\documentclass[9pt, sigconf, nonacm]{acmart}

\ifluatex
\else
\DeclareUnicodeCharacter{2212}{−}
\fi

\usepackage[]{cleveref} 
\usepackage{multirow}
\usepackage{dynamicwidth}
\usepackage{subcaption}
\usepackage{circledsteps}
\usepackage[frozencache,cachedir=.]{minted}
\definecolor{LightGray}{gray}{0.95}

\newcommand*{\tensorsockets}{\texorpdfstring{{\textsc{TensorSocket }}}{TensorSocket}} 
\newcommand*{\tensorsocket}{\texorpdfstring{{\textsc{TensorSocket}}}{TensorSocket}}

\AtBeginDocument{%
  }

\begin{document}

\title{TensorSocket: Shared Data Loading for Deep Learning Training}

\author{Ties Robroek}
\affiliation{%
  \institution{IT University of Copenhagen}
  \country{}
  \city{}
}
\email{titr@itu.dk}

\author{Neil Kim Nielsen}
\affiliation{%
  \institution{IT University of Copenhagen}
  \country{}
  \city{}
}
\email{neilkimn@protonmail.com}

\author{Pınar Tözün}
\affiliation{%
  \institution{IT University of Copenhagen}
  \country{}
  \city{}
}
\email{pito@itu.dk}

\renewcommand{\shortauthors}{Robroek et al.}

\input{sections/0-abstract}

\maketitle
\input{sections/1-introduction}
\input{sections/2-background}
\input{sections/3-implementation}
\input{sections/4-results}
\input{sections/5-discussion}
\input{sections/6-related-work}
\input{sections/7-conclusion}

\section*{Acknowledgements}

This work is funded by 
the Independent Research Fund Denmark's (Danmarks Frie Forskningsfond; DFF) Sapere Aude and Inge Leh-mann programs under grant agreement number 0171-00061B and 0171-00062B, respectively.
We also thank DSAR lab members at IT University of Copenhagen for their support.

\bibliographystyle{ACM-Reference-Format}
\bibliography{biblio}

\end{document}

%% file: sections/0-abstract.tex
\begin{abstract}

Training deep learning models is a repetitive and resource-intensive process.
Data scientists often train several models before landing on a set of parameters (e.g., hyper-parameter tuning) and model architecture (e.g., neural architecture search), among other things that yield the highest accuracy.
The computational efficiency of these training tasks depends highly on 
how well the training data is supplied to the training process.
The repetitive nature of these tasks results in the same data processing pipelines running over and over, exacerbating the need for and costs of computational resources.

In this paper, we present \tensorsockets to reduce the computational needs of deep learning training by enabling simultaneous training processes to share the same data loader.
\tensorsockets mitigates CPU-side bottlenecks in cases where the collocated training workloads have high throughput on GPU, but are held back by lower data-loading throughput on CPU.
\tensorsockets achieves this by reducing redundant computations and data duplication across collocated training processes and leveraging modern GPU-GPU interconnects. 
While doing so, \tensorsockets
is able to train and balance differently-sized models and serve multiple batch sizes simultaneously
and is hardware- and pipeline-agnostic in nature.

Our evaluation shows that
\tensorsockets enables scenarios that are infeasible without data sharing, increases training throughput by up to $100\%$,
and when utilizing cloud instances, achieves cost savings of $50\%$ by reducing the hardware resource needs on the CPU side.
Furthermore,  
\tensorsockets outperforms the state-of-the-art solutions for shared data loading
such as CoorDL and Joader;
it is easier to deploy and maintain
and either achieves higher or matches their throughput
while requiring fewer CPU resources.

\end{abstract}

%% file: sections/1-introduction.tex
\begin{figure}[h]
\centering
\includegraphics[width=0.99\linewidth]{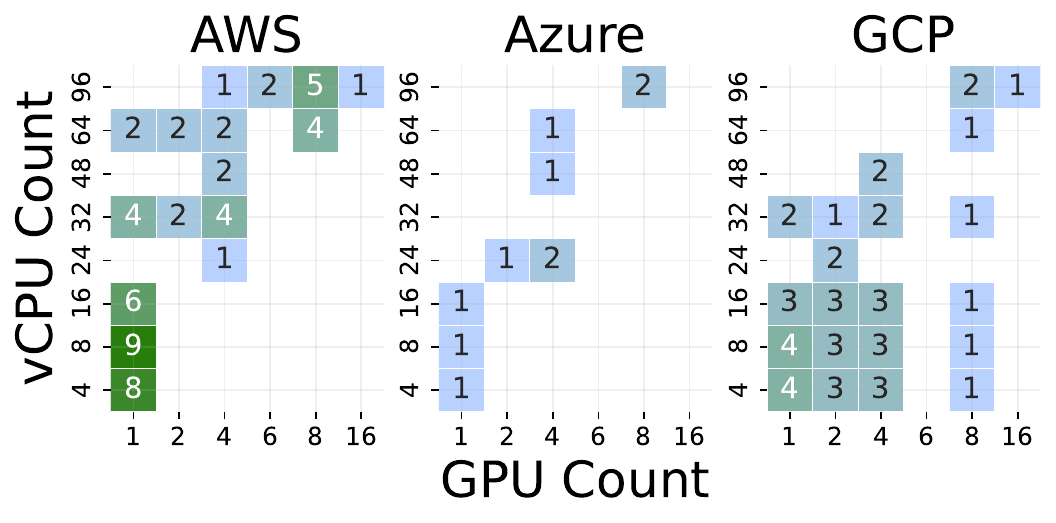}
    \caption{Cloud instances by vCPU to GPU ratio of Amazon Web Services (AWS) \cite{aws_ratio}, Microsoft Azure \cite{azure_ratio}, \& Google Cloud Platform (GCP) \cite{gcp_ratio}.
    The color scale denotes the number of instances offered with the specified vCPU-GPU pairing.}
\label{fig:vcpu_gpu_ratio}
\Description[Cloud instances by vCPU to GPU ratio of popular cloud providers.]{Cloud instances by vCPU to GPU ratio of Amazon Web Services (AWS), Microsoft Azure, \& Google Cloud Platform (GCP).}
\end{figure}

\section{Introduction}
\label{sec:introduction}

The process of training a deep learning (DL) model is computationally expensive, mandating the use of powerful accelerators such as GPUs to match the computational needs.
However, while the core of the training process can naturally be accelerated this way for many DL models,
some training pipelines feature computationally expensive data pre-processing operations such as augmentation and decoding \cite{cubuk2019autoaugment}.
Such operations often cause bottlenecks on the host-, or input-side of the training pipeline,
where the dominating processing unit is still the CPU \cite{cachew, isenko}.

Compute offerings of cloud providers are popular for addressing the computational needs of deep learning training thanks to their on-demand availability.
On the other hand, the range of CPU-to-GPU configurations is rather limited, as shown in \Cref{fig:vcpu_gpu_ratio}.
Furthermore, an instance with a high vCPU to GPU ratio can cost up to $16$ times as much as an instance with minimal vCPU count with the same GPU \cite{aws_ratio}.
This high trade-off for the need for more CPU availability, combined with the wide range of DL workloads and their differing computational requirements,
lead to several bottlenecks \cite{tf.data, mohan2021analyzing, 10.1145/3592980.3595314}.
Specifically, DL training processes that are bottlenecked by their input processing 
render expensive high-performance accelerators underutilized
\cite{microsoftDLClusterStudy, alibabaStudyGPU}.
In turn, this wastes both CPU and GPU resources.
Underutilization of cloud resources is financially wasteful for everyone as compute that has been paid for is not used effectively.
Furthermore, it creates an unsustainable carbon footprint in order to address the demand for AI \cite{carbonIntensity, greenai, StrubellGM19, googleCarbon, carbontracker, hypesustainability}.

\begin{figure*}
\centering
\begin{subfigure}{.495\textwidth}
\centering
\includegraphics[width=0.99\linewidth]{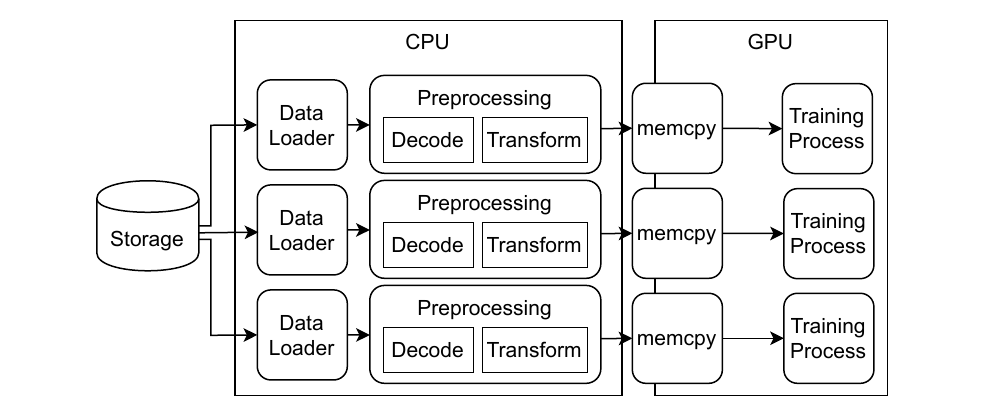}
\caption{Conventional DL input pipeline.}
\label{fig:single_loading}
\Description[Conventional DL input pipeline.]{Conventional DL input pipeline with redundancies between training processes.}
\end{subfigure}
\begin{subfigure}{.495\textwidth}
\centering
\includegraphics[width=0.99\linewidth]{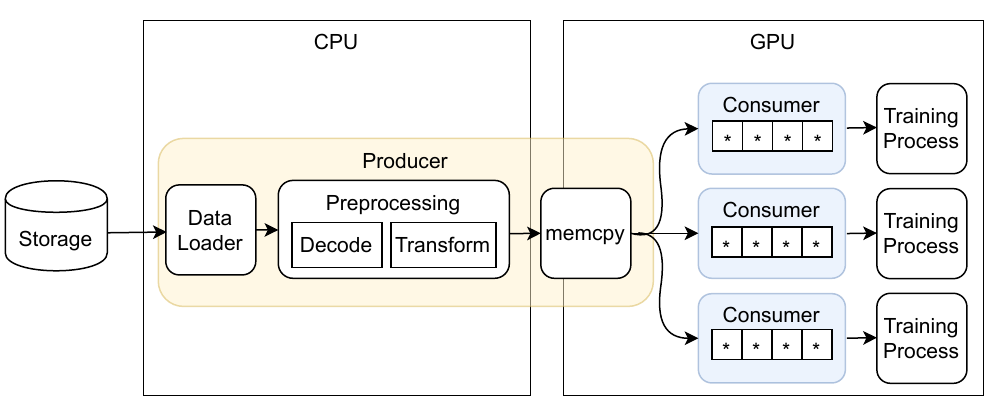}
\caption{Shared input pipeline with \tensorsocket.}
\label{fig:shared_loader}
\Description[Shared input pipeline with \tensorsocket.]{Shared input pipeline with \tensorsocket without data redundancies.}
\end{subfigure}
\caption{Collocated DL model training (a) without sharing the data loader and (b) with sharing the data loader using \tensorsocket.  The arrows denote the flow of data from storage to training processes. The shared data loading process (producer) is shown in yellow, and the training processes (consumers) in blue.}
\label{fig:data_loader}
\end{figure*}

In practice, it is common to train several models to accomplish a task.
The feasibility of DL models heavily depends on finding a model architecture (neural architecture search) or a set of parameters (hyper-parameter tuning) that responds well to the data.
This results in model training scenarios that exhibit shared tasks, especially in their input pipelines.

There has been recent work that has proposed ways to leverage such shared tasks
\cite{mohan2021analyzing, audibert2023tf, 234827, xuDeepLearningDataloader2022}
and effectively demonstrated the premise of sharing.
On the other hand, these works either
focus on the cloud scale, paying less attention to the finer-grained cooperation across training processes on the same server,
or put a heavy burden on the CPU resources, essentially locking the CPU into being a data feeder for a hardware accelerator instead of facilitating efficient resource utilization.

In this paper,
we draw inspiration from these prior works on data sharing across DL training tasks
in addition to the work on database systems that leverage the shared work done across concurrent database requests \cite{qpipe, shareddb, addict, psaroudakisAA13}.
Our goal is to increase opportunities for work sharing and collocation across DL training jobs
while minimizing the hardware resource requirements for such jobs.
Rather than viewing these jobs as big monolithic isolated tasks that have to get scheduled exclusively on some CPU and GPU resources, we propose \tensorsocket,
a novel data loader that is shared across models being trained on the same dataset. 
\tensorsockets turns the \textit{competition} for data and hardware resources into a \textit{cooperation}
allowing for more effective workload collocation across concurrent training tasks.
As a result, it alleviates resource underutilization and the aggregate costs of DL model training.

Our contributions are as follows:
\begin{list}{\labelitemi}{\leftmargin=1.5em}
\item{We present the design and implementation of a novel data loader, \tensorsocket,
that enables work and data sharing across collocated models training on the same dataset
by detaching the data loading pipeline from the training process.
\tensorsockets employs techniques such as \textit{batch buffering} and \textit{flexible batch sizing}
to support models joining at different times, models of varying complexity, and different batch sizes and order across models as the collocated training process continues.}
\item{We demonstrate how to adopt \tensorsockets as an alternative data loader in training pipelines implemented in PyTorch \cite{pytorch} in a plug-and-play manner
and evaluate its benefits across a variety of training scenarios (image analysis, audio classification, generative AI) and hardware setups (powerful on-premises hardware, different cloud offerings).
Our evaluation shows that \tensorsockets enables scenarios that are infeasible without data sharing,
doubles training throughput for some scenarios,
and, most importantly, when utilizing cloud instances,
can halve the cloud costs by reducing the CPU needs by up to four times.}
\item{We compare \tensorsockets to the state-of-the-art techniques for shared data loading,
CoorDL \cite{mohan2021analyzing} and Joader \cite{xuDeepLearningDataloader2022},
and demonstrate that \tensorsockets either outperforms or matches their training throughput
while requiring less CPU resources and lower deployment effort.}
\end{list}

The rest of this paper is structured as follows.
\Cref{sec:background} gives an overview of data loading in DL model training,
before \Cref{sec:implementation} presents \tensorsockets and motivates concrete use case scenarios for it.
Then, \Cref{sec:results} demonstrates the multi-dimensional benefits of \tensorsockets over a variety of training pipelines and hardware setups,
and \Cref{sec:vision} discusses its further applicability.
Finally, \Cref{sec:related_work} surveys related work,
and \Cref{sec:conclusion} concludes the paper.

%% file: sections/2-background.tex
\section{Data Loading in Deep Learning}
\label{sec:background}

\textbf{Characteristics and bottlenecks.} \Cref{fig:single_loading} shows the different training processes that deal with reading and transforming data.
In general, a DL training process consists of a model configuration,
a dataset partitioned for training and validation,
and a training loop.
The training process iterates over the full training partition of the dataset for a number of times, specified as \textit{epochs}.
The \textit{data loader} is tasked with fetching and readying the data for the model to train on.
During each iteration, a \textit{batch} of data is prepared from either disk or memory.
This preparation includes fetching, decoding, and transforming the samples.
Furthermore, data may be augmented in order to improve the accuracy of models trained on it and their ability to generalize to future unseen data \cite{cubuk2019autoaugment, VenturaKQM21}.

Decoding, transforming, and augmenting data are all steps that handle and modify data during training and are collectively called \textit{data pre-processing} operations. The more extensive the pre-processing, the higher the computational overhead during training, potentially introducing input-bound bottlenecks. In some real-world training, these operations can amount to half of the energy costs \cite{metaDataLoadingAnalysis}.

Furthermore, even though the data used for training is in fast local storage, such as main-memory or SSDs, this data commonly exceeds memory capacity. The result is that the data has to be repeatedly read from disk, swapping out the data that has already been trained on.
The damage done by this swapping and OS thrashing depends on how much data can fit in memory as well as the storage backend. This introduces I/O as a potential bottleneck on the critical path \cite{mohan2021analyzing}.

\textbf{Alleviating the bottlenecks.}
Data loaders can be configured to alleviate problems that may arise due to inadequate host-side resources that result in the training process idling the GPU \cite{tf.data}.
For example, pre-fetching overlaps the work done for data preparation with model training by reading and processing data prior to when it is required during training. Similarly, scaling up the number of workers contributing to reading and pre-processing the data can help hide the input bottlenecks in training. 
While increasing the worker count does not speed up the pre-processing time of individual batches, it does increase the total batch throughput that can be fed to the model training.
On the other hand,
a high degree of pre-fetching and parallel workers incur higher host-side CPU utilization and memory consumption, increasing the resource cost and potentially causing contention for resources at the host side.

If the CPU is fully utilized while the GPU is not, another option is to offload pre-processing operations to the GPU. Tools like NVIDIA DALI \cite{nvidia_dali} or techniques like FusionFlow \cite{fusionflow} and FastFlow \cite{fastflow} offer methods for resolving data loading bottlenecks that may arise in such scenarios. 
However, offloading pre-processing to the GPU reserves compute resources that could otherwise be used for training the model itself and should therefore be done with care.

\textbf{Opportunities for sharing.}
Developing an effective DL model often requires multiple models to be trained and evaluated in quick succession over the same or similar datasets.
Model selection is a common practice where model architectures and training configurations are empirically compared.
Hyper-parameter tuning evaluates different hyper-parameter settings, such as learning rate, weight decay, and optimizer settings.
These types of tasks are essential for landing on the best performing model ~\cite{probst2018tunability, li2020massively, liu2018progressive, cerebro}, as the range of different model architectures and hyper-parameters available increase with the introduction of new models.

Our goal is to propose a mechanism to alleviate the data loading bottlenecks in deep learning training that is complementary to the ones listed above.
More specifically,
motivated by the repetitive nature of tasks during the model search and hyper-parameter tuning for deep learning training,
we would like to leverage the shared data and work required by these tasks.
In our solution, \tensorsocket,
we aim to dedicate the maximum amount of GPU resources to the training loop itself
and minimize CPU resource requirements for data loading.
Furthermore, by extending existing data loader implementations instead of replacing them, our solution is compatible with other optimizations that can be done for data pre-processing.

\textbf{State-of-the-art sharing techniques.}
Motivated by these opportunities for sharing,
prior work has also advocated for data sharing in DL
\cite{mohan2021analyzing, audibert2023tf, 234827, xuDeepLearningDataloader2022}.
Here, we more specifically detail the proposals that are closest to \tensorsocket,
which we also compare \tensorsockets against in \Cref{sec:results}.

\textit{CoorDL} \cite{mohan2021analyzing} is an extension to NVIDIA DALI that coordinates data pre-processing. It is designed for the cluster level and can be used to share data directly between training processes. It can distribute a batch of data to any number of training processes in the cluster. Once all training processes are done with the data, CoorDL continues to the next batch. CoorDL's focus on the cluster level and design around DALI, however, surfaces some limitations. Firstly, CoorDL is designed for models training on separate GPUs and cannot utilize leftover GPU compute power to train multiple models on a single GPU in a collocated fashion.
Secondly, CoorDL performs poorly when the models that train simultaneously are not very similar, as in this case the models that are faster have to wait for the slower ones. 
This rigid design also prevents CoorDL from being deployed as a live service with training processes arriving at different moments.
Thirdly, CoorDL requires the data loading and pre-processing pipeline to be implemented in DALI, requiring substantial extra engineering if the pipeline implementation is not already using DALI.
Finally, the existing CoorDL codebase \cite{coordlcodebase}, and the DS-Analyzer project around it, is written in Python 3.6, which has been deprecated since 2021 by PyTorch and reached Python end-of-life in December that year.

\textit{Joader} \cite{xuDeepLearningDataloader2022} is a standalone shared data loading solution that supports sharing over multiple datasets. A server is configured in which all datasets have been registered. Training clients, also known as jobs, then communicate with this server using RPC. Joader reduces CPU utilization by having one server that does the data loading and pre-processing for multiple jobs, even if those jobs require datasets that are not identical but just overlap. In the scenario where models train on the same base dataset, this allows models to train at different speeds and still share part of their data loading.
Joader achieves this flexibility across different datasets through a technique called dependent sampling. 
However, this sampling also comes with an important drawback;
it requires intersection calculations to run at every iteration, which adds a high CPU cost.
Furthermore, Joader only has a proof-of-concept implementation \cite{joadercodebase}, which is written in Rust making it very difficult to adapt existing deep learning training codebases to. Datasets have to be converted to the proprietary format that Joader expects and only image data with specific parameterization is supported. There is no support for a variety of image pre-processing operations other than the pipeline that is hardcoded in Rust.
Finally, data reaches the training jobs as NumPy matrices which require tensor conversion and host-to-device transfer, and batching during training is not supported,
which are all detrimental to data loading and training performance.

Next, we delve deeper into \tensorsocket.

%% file: sections/3-implementation.tex
\section{\tensorsocket}
\label{sec:implementation}

This section presents \tensorsocket\footnote{\url{https://github.com/itu-rad/tensorsocket}}, our shared data loader that capitalizes on the redundancy among similar but separate data loaders of collocated training processes.
We propose a solution to inefficient hardware utilization and resource wastage by minimizing redundant work and hardware resource consumption while ensuring that downstream training processes are not impacted.
By detaching the data loading pipeline from each training process, we can merge several training processes into a single data loading pipeline. This single data loader can expose the training data for use in each collocated training process.

\subsection{Overview}
\label{sec:implementation:overview}

\Cref{fig:shared_loader} illustrates how our shared data loader works.
The figure shows an example of three collocated training processes, where, in yellow, the tasks of the \textit{producer} are shown, and in blue, the training process with the \textit{consumer} is shown. 

At its core, our system is composed of a \textit{producer} and a number of \textit{consumers}. The \textit{producer} holds a single data loading process along with some bookkeeping, while the \textit{consumers} iterate on data sent by the producer. This producer-consumer workflow can be swapped in place of DL training framework-specific \texttt{DataLoader} objects, such as the PyTorch DataLoader.

Given that the producer is the owner of the data loading pipeline as well as responsible for data generation, it would be regressive to copy each batch of data into every collocated training process. Instead, once the producer has prepared a batch of data, the consumers are all given the location of the data batch to use in their respective processes. Every consumer has a queue that holds up to a few of these locations. This introduces some flexibility to prevent training hiccups (e.g., a training process falling behind during a batch) from interfering with the other training processes.

\begin{figure}[t]
\begin{subfigure}{.495\textwidth}
\begin{minted}
[
frame=lines,
framesep=2mm,
baselinestretch=0.8,
bgcolor=LightGray,
fontsize=\footnotesize,
linenos
]
{python}
# train.py (without TensorSocket)
data_loader = DataLoader(dataset)
for batch in data_loader:  # Loop over dataset
    ...  # Model training iteration
\end{minted}
\caption{Conventional training script without \tensorsocket.}
\label{listing:implementation-example-none}
\end{subfigure}
\huge{$\downarrow$}
\begin{subfigure}{.495\textwidth}
\begin{minted}
[
frame=lines,
framesep=2mm,
baselinestretch=0.8,
bgcolor=LightGray,
fontsize=\footnotesize,
linenos
]
{python}
# producer.py
data_loader = DataLoader(dataset)
producer = TensorProducer(data_loader)
for _ in producer:  # Loop over dataset, send data
    pass
producer.join()  # Clean-up
\end{minted}
\caption{\tensorsockets producer script.}
\label{listing:implementation-example-producer}
\end{subfigure}
\begin{subfigure}{.495\textwidth}
\begin{minted}
[
frame=lines,
framesep=2mm,
baselinestretch=0.8,
bgcolor=LightGray,
fontsize=\footnotesize,
linenos
]
{python}
# consumer.py (or train.py)
consumer = TensorConsumer()
for batch in consumer:  # Receive & loop over data
    ...  # Model training iteration
\end{minted}
\caption{\tensorsockets consumer script.}
\label{listing:implementation-example-consumer}
\end{subfigure}
\caption{Example \tensorsockets implementation requiring minimal code changes training for a single epoch. The top listing depicts a standard model training script. The data loader is split from the main training process, creating a producer process and a consumer process.}
\label{listing:implementation-example}
\Description[Example \tensorsockets implementation.]{Example \tensorsockets implementation showing how to adapt an existing codebase to \tensorsocket.}
\end{figure}

\subsection{Design \& Implementation}
\label{sec:implementation:implementation}

\tensorsockets is a library built around PyTorch as it is the most widely used deep learning framework available. A crucial limitation common in other data sharing solutions \cite{mohan2021analyzing, xuDeepLearningDataloader2022} discussed in \Cref{sec:background} is that the solution itself implements the complete data loader. This limits the adoption as it requires the user to adapt to the specific codebase, in addition to the library version dependencies, of that solution. We prevent these shortcomings by setting \tensorsockets up as a wrapper around data loaders implemented in PyTorch instead of a separate data loader itself. This ensures out-of-the-box compatibility with any PyTorch training script, e.g. ones that use PyTorch or Hugging Face data loaders.
While this means that the current implementation is PyTorch specific,
implementations of similar wrappers for other frameworks such as TensorFlow 
won't be prohibitive as these frameworks generally follow the same principles for data loading.

\Cref{listing:implementation-example} shows an example usage of \tensorsocket.
Using \tensorsockets involves copying the data loading logic of a training script to a separate producer script and adding a consumer to the training script.\footnote{The full example can be found in our code repository at \url{https://github.com/itu-rad/data-sharing}.}The PyTorch \textit{data loader} is isolated to a different process and wrapped in a \textit{TensorProducer}, which is then iterated over similar to a conventional data loader.
The training process receives the data automatically by iterating over a \textit{TensorConsumer}.

\begin{figure}
    \includegraphics[width=0.99\linewidth]{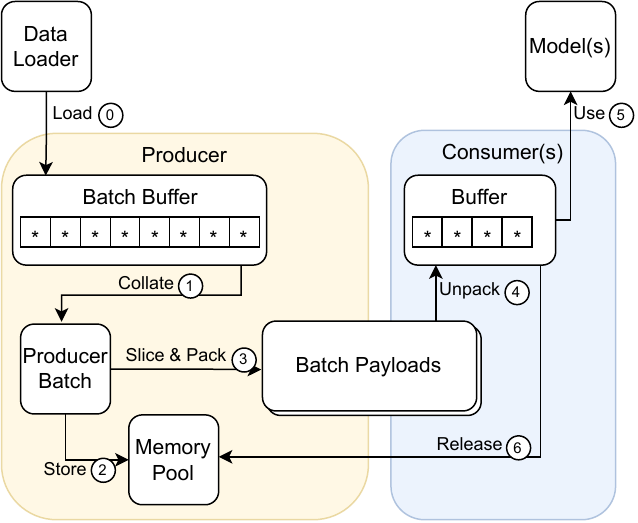}
    \caption{Overview of \tensorsocket's internals. Arrows denote PyTorch tensor and pointer operations.}
    \label{fig:pytorchimpl}
    \Description[Diagram of \tensorsocket's internals.]{Diagram of \tensorsocket's internals depicting all of the data movements.}
\end{figure}

\Cref{fig:pytorchimpl} illustrates \tensorsocket's key components and operations,
which we detail next.

\subsubsection{Producer}
\tensorsockets splits data loading from training.
The data loading producer becomes a server that can dynamically process and serve incoming consumer clients.
A \tensorsockets producer instance is initialized with a data loader object. It is exposed as an iterator that itself iterates over the nested data loader it is initialized with. The producer repeatedly requests~\Circled{0} the contained data loader to fetch the data from disk. It will pause iterating over the data loader whenever the consumers currently do not need extra data, notified by communication between the producer and consumers. This can also be the case when there are no consumers present as in this case there is no need for any data loading. 

\subsubsection{Consumer}
Abstracting away the data loader into the producer allows the consumer to be lightweight. As \tensorsocket's producer and consumer directly replace the data loading batch iterator in the training script, the consumer similarly takes the form of an iterable object that fetches new data~\Circled{5} whenever available. If there is no data available, the consumer halts and waits. This design makes for a minimal one-line swap in training script code.

\subsubsection{Communication}
The communication between the producer and consumers is done using ZeroMQ sockets. ZeroMQ is an open-source library that allows for sharing atomic messages with low latency. In \tensorsocket's case, we use ZeroMQ sockets for communication between the producer and consumers using a PUB/SUB pattern \cite{zeromq}. This is a multicast pattern that is flexible and scalable, allowing one producer to connect to multiple consumers.

The data is shared over these sockets by the producer. Once a consumer has fetched a readied batch of data for use in training, it notifies the producer by sending an acknowledgment message back to it, allowing for the producer to continuously keep the consumers fed new data. When multiple consumers are training simultaneously, the producer will wait for an acknowledgment from all consumers before releasing a piece of data. This ensures that all consumers have iterated on a batch before it is deleted.

Whenever data is shared with a consumer, the producer will store~\Circled{2} a reference to that data. Once a consumer has finished a batch and moves on to the next, it will notify~\Circled{6} the producer. The producer will release the associated memory when all consumers are finished with it, allowing the producer to ready the next batch.

Depending on the dataset and models, consumers may take a long time to go through their training data batches. In order to be continuously aware of consumers, producers send and receive heartbeat messages from their consumers over a different socket. The producer will detach from consumers that it has not received a heartbeat from in a while.

\subsubsection{Data sharing}
\label{sec:implementation:sharing}
Data sharing is at the heart of \tensorsocket.
If the data sharing implementation is not efficient, it will become a bottleneck that would outweigh the benefits of sharing.
There are two ways data sharing can be done. 

Some solutions share the data bytes directly with the training processes via inter-process communication \cite{xuDeepLearningDataloader2022}. This surfaces some concerns regarding sharing efficiency and data duplication. Increasing the size of the training data directly increases the size of the network messages in such a solution, potentially leading to slowdowns. Furthermore, while the data loading itself is unified, the resulting data is then duplicated for every client, spiking memory consumption and data movement costs.

In \tensorsocket's case, we share small packets containing pointers to the data instead of the data itself. Following our earlier design philosophy, we heavily borrow from PyTorch's
existing data management. 
PyTorch introduces Tensor objects which are data matrices, similar to NumPy matrices, that contain all data that PyTorch runs on. While PyTorch is most commonly known as a Python library, much of the internals such as Tensors are defined in C++. We can use this to our advantage by extracting the data pointer and other necessary information. This pointer is then shared by the producer to the consumers, which in turn use this data to reconstruct the Python tensor object without any data duplication.

PyTorch as a library is heavily optimized for running with multiple threads as data workers and on multiple GPUs and machines. The tensor implementation contains methods for dealing with concurrency and distribution, including tensor rebuilding. By using this somewhat hidden PyTorch functionality we can pack~\Circled{3} and unpack~\Circled{4} batch payloads of tensors.

Tensors in PyTorch hold data that can be on the host system (i.e., CPU), but can also be put on the GPU. Transferring data to the GPU is a costly operation. By inheriting PyTorch tensor methods, \tensorsockets can reconstruct tensors on both the CPU and GPU. This means that the producer can put the data on the GPU once, after which all consumers collocated on that GPU can access it. Furthermore, we can rely on PyTorch's tensor management for our shared data. Tensors are kept in memory as long as any of the producers or consumers hold a reference to it. 

Finally, using the capabilities of frameworks such as PyTorch gives us access to fast GPU-to-GPU communication methods like NVLink while sharing the tensors. This allows \tensorsockets to efficiently share data even if the models train on different GPUs. Data is loaded on one of the GPUs after which it is directly shared to the other GPUs with direct NVLink interconnects.

\subsubsection{Synchronization}
\label{sec:implementation:sync}
Considering that consumers may train different models and training is stochastic, we should expect that consumers do not process a batch in identical time. This led us to introduce a batch buffer on the consumer side. Instead of actively requesting the next batch on iteration, consumers can hold up to N batches (i.e., pointers to the tensors of batches) in their buffer. This allows for the producer to actively pre-fetch data, and for the consumers to drift at most N batches apart. Both the buffering and the pre-fetching hide the latency of various parts of the data loading pipeline. When designing this queue, we experimentally found that a buffer as small as two batches is enough to provide maximum training throughput while training similar tasks. Increasing the buffer size can be beneficial when training processes fluctuate more widely in their speed. It should be noted that increasing the buffer size does increase the GPU memory requirement of the system as more batches need to be kept on the GPU simultaneously.

\begin{figure*}[t!]
\centering
\includegraphics[width=1\linewidth, clip]{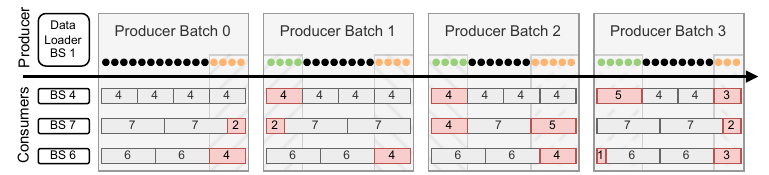}
    \caption{\tensorsockets with flexible batch sizing training consumers with different batch sizes. The batch sizes have been divided by their common multiple for clarity. The producer collates 16 data loader batches into a single producer batch and then splits the producer batch into appropriate batches per consumer. These consumers request batches of 4, 7, and 6 respectively. These batch sizes do not all divide 16 cleanly, which results in a fraction of the data having to be repeated. The repeated amount of data points can vary between batches; batch 1 and 2 repeat four while batch 3 repeats five.}
\label{fig:flexible-batches}
\Description[\tensorsocket's flexible batch sizing.]{\tensorsocket's flexible batch sizing. When batch sizes do not divide equally, part of the previous batch has to be repeated.}
\end{figure*}

While the buffering scheme described above relaxes the conditions for data sharing among the consumers, \tensorsockets by design targets scenarios where the consumers train on the same dataset at or around the same time.
Therefore, the consumers have to be balanced in terms of their training speed even if they do not process data in identical time.
Whenever a process trains too fast, the consumer iterator will automatically halt as there is no data available. 
This frees up resources for other consumers to make up for the difference.
In this case, the GPU can be time-shared among the consumers. 
Modern GPUs enable this through services such as NVIDIA multi-streams or Multi-Process Service (MPS) \cite{robroek2023analysis}.
Especially, MPS allows efficient time and spatial sharing of the GPU.
The GPU sharing and inclusion of the consumer buffer allow to balance the load of the consumers automatically, resulting in higher training throughput and GPU utilization. As a result, \tensorsockets is able to train models efficiently, even when models diverge in terms of complexity. GPU resources are allocated in such a way that the consumers go through the data at the same rate.

Since the \tensorsockets producer acts as a server producing data for the consumers,
we also need to account for consumers that connect at different times.
Once an epoch has already started, any new consumers fall behind and have to wait for the next epoch to start training. We introduce a leniency measure called rubberbanding to provide a window for consumers to join training. If a consumer joins before 2\% of the dataset has been iterated on in an epoch, the producer will halt all other consumers to let that consumer synchronize. The percentage of the dataset that serves as the cutoff point is configurable. We found that rubberbanding is an effective method for allowing users to spawn multiple consumers without fear of them not joining fast enough.

\subsubsection{Flexible Batch Sizing}
By default, \tensorsockets restricts consumers to training on the same batches at the same time, requiring all training processes to utilize the same batch size. 
We introduce an alternative mode of operation to relax these constraints. Under \textit{flexible batch sizing}, consumers tell the producer what batch size the producer should supply them. The producer supplies batches of different sizes to their respective consumers using the slicing properties of PyTorch tensors. Enabling flexible batch sizing is done by setting batch size arguments when initializing the producer and the consumers, lines 2 in Figures~\ref{listing:implementation-example-producer} and \ref{listing:implementation-example-consumer}.

\tensorsockets directly share the pointers to the batch tensors under default operation. With flexible batches, the producer and consumer batches are no longer the same. Instead, the producer readies \textit{producer batches}, which are larger than the batch sizes of the consumer. The producer batch is allocated as a continuous block of memory on the GPU. Pointers to appropriate sliced batches are sent to the consumers. This way, the consumers can train on varying batch sizes, while still going through the data at the same rate.

\Cref{fig:flexible-batches} illustrates flexible batch sizing in practice. In this example, the producer serves consumers with requested batch sizes of 4, 7, and 6. The producer collates~\Circled{1} the data it receives from the data loader into producer batch sizes of 16. These batches are then sliced~\Circled{3} appropriately for every consumer.

When the consumer batch sizes do not cleanly divide the producer batch size, a share of data has to be repeated between producer batches.
The amount of repeated data for a producer batch can be up to (not including) the largest consumer batch size $\max\{b_c\}-1$, as otherwise more consumer batches would have fit in that producer batch. The maximum repeated share divides this by the size of the producer batch $\frac{\max\{b_c\}-1}{b_p}$. As a large producer batch minimizes repetition, we recommend having it at least twice as large as the largest consumer batch, making this share never exceed $50\%$.

We argue that data repetitions are an acceptable tradeoff for flexibility as \tensorsockets reduces data movements considerably w/o flexible batching enabled. Repetitions only occur if the consumer batch sizes do not cleanly divide the producer batch size, with \Cref{fig:flexible-batches} depicting a challenging scenario. In practice, we expect the batch sizes requested by consumers to 
be in powers of two.

\subsubsection{Batch Order}
\tensorsocket's unified data loading results in consumers training on the data points in the same sequence. For some tasks, such as hyperparameter tuning or model architecture search, it may be beneficial to include more variation by loosening this order requirement.

With producer batches, we introduce two ways to add variety in how the consumers go through the data. Firstly, we can utilize the flexible batch sizing mode and add offsets to the consumers, causing their batches to be carved from the producer batch with different breakpoints. This results in batches that are not identical between consumers. Secondly, we can shuffle the order of batches within a producer batch. If a consumer splits a producer batch in, e.g., 4 batches, it can then go through these in random order. This results in data being visited in different order between consumers. If desired, both techniques can be combined. The amount of variation in visiting then becomes a product of producer batch size, at the cost of GPU memory. Similar to flexible batch sizes, these two techniques can be activated when initializing the producer.

\begin{figure*}[t!]
\centering
\includegraphics[width=0.99\linewidth]{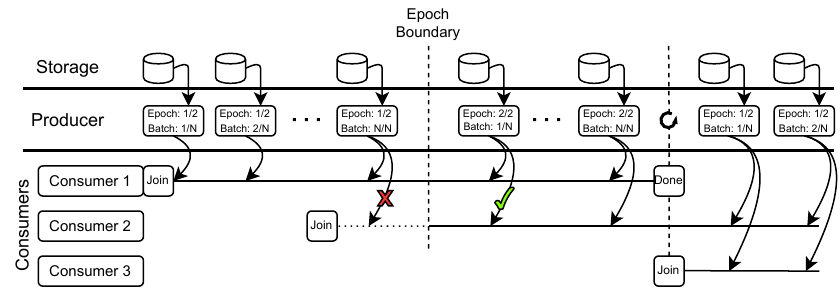}
    \caption{Illustration of how \tensorsockets allows the producer to run continuously and supply new consumers with batches.}
\label{fig:indefinite_producer}
\Description[\tensorsockets running continuously on a server.]{\tensorsockets running continuously on a server. The producer accepts any incoming consumers to train models on the dataset.}
\end{figure*}

\subsection{Use Case Scenarios}
\label{sec:implementation:scenarios}

We go over a few use case scenarios that would benefit from \tensorsocket, and how our implementation makes them possible.

\subsubsection{Centralized Always-Available Loading.}
\label{sec:implementation:scenarios:centralized}

When exploring a dataset, it is invaluable for users to seamlessly start and stop training jobs.
\tensorsockets allows for a high degree of flexibility and stability by abstracting away the data loading from the training job itself. Once \tensorsockets is running on a server, consumers can come and go as they please. The consumers ping the producer with heartbeats such that the producer knows how many training processes are active at a given time. Consumers may either join training at any point in an epoch or wait until a new epoch starts, depending on the configuration.
In the latter case, the producer buffers the first few batches at the start of a new epoch to provide a short window in which new consumers are accepted. In such a case the producer will halt the other consumers in order for the new consumer to
quickly iterate over the buffer.

\Cref{fig:indefinite_producer} shows an example of how new jobs, represented as consumers, are handled by our shared data loading system. In the example, consumer 2 joins in too late during the first epoch, having to wait until the second epoch starts. Consumer 3 joins in at an epoch boundary and immediately starts consuming data batches.

\subsubsection{Native Inter- and Intra-GPU Sharing.}
\label{sec:implementation:scenarios:collocation}

One way of increasing the hardware utilization for DL training is training multiple models at the same time, i.e., \textit{workload collocation} \cite{DelimitrouK14, dicer}.
Workload collocation can improve training throughput \cite{hfta, robroek2023analysis, orion, 10.1145/3642970.3655830}
when the hardware resource needs of the individual training processes
are not large enough to utilize all the available CPU and GPU resources \cite{crossbow, baunsgaardWT20} or are bottlenecked by their input pipelines \cite{behme_art_2023}.
This often reduces the aggregate runtime when more than one model has to be trained,
even though the training time per model usually goes up due to not having exclusive access to the GPU.

\tensorsockets allows for sharing data on a single GPU between any number of consumers,
boosting efficiency while collocating models.
This additionally reduces redundant memory consumption, as the memory requirement for training processes is lowered due to not needing a data loader for each separate training process. 

Our implementation also supports collocation across multiple GPUs. Data batches are seamlessly moved between GPUs when the data is needed on a different device. Thus, \tensorsockets is able to leverage GPU-GPU interconnects with lower latency and higher bandwidth than CPU-GPU interconnects, as also mentioned in \Cref{sec:implementation:sharing}.
We evaluate this scenario in \Cref{sec:results:image}.

\subsubsection{Sharing for Mixed Workloads.}
\label{sec:implementation:scenarios:mixed}

Mixed workloads that train models at different speeds, for instance when model complexity differs significantly, can be difficult to optimize from a data loading perspective.
\tensorsockets supports mixed workloads by allocating more hardware resources to heavier training processes than lighter processes. We bound the models to be within a certain amount of batches from each other (as described in \Cref{sec:implementation:sync}).
The result is that slower models are sped up and lighter models are slowed down so that all models traverse the epoch in the same amount of time.
We evaluate this scenario in \Cref{sec:results:mixed}.

\subsubsection{Sharing Generative Tasks Online.}
\label{sec:implementation:scenarios:generative}

\begin{figure*}
\centering
\begin{subfigure}{.495\textwidth}
\centering
\includegraphics[width=0.99\linewidth]{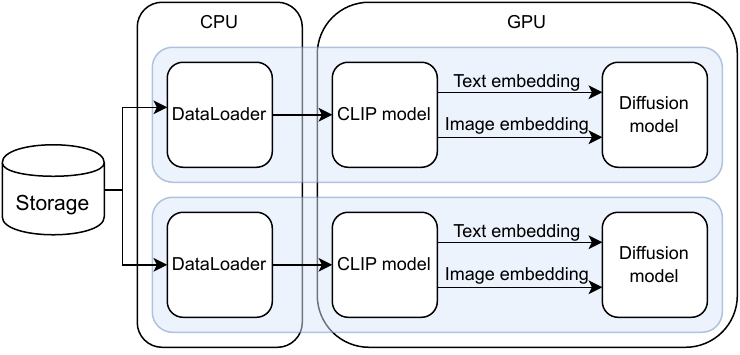}
\caption{Conventional deep learning input pipeline.}
\label{fig:dalle_single_loading}
\Description[Conventional deep learning input pipeline using a CLIP model on the gpu.]{Conventional deep learning input pipeline using a CLIP model on the gpu.}
\end{subfigure}
\begin{subfigure}{.495\textwidth}
\centering
\includegraphics[width=0.99\linewidth]{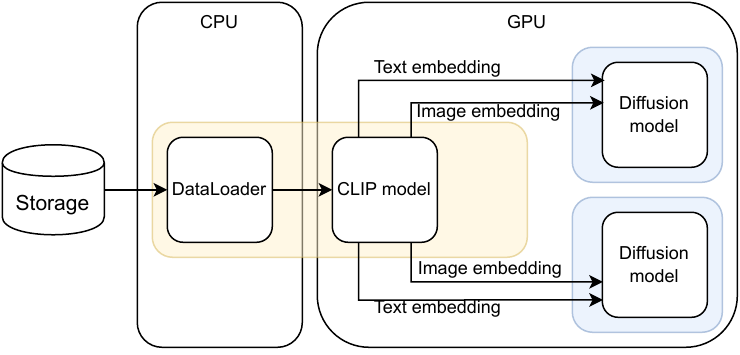}
\caption{Shared input pipeline with \tensorsocket.}
\label{fig:dalle_shared_loader}
\Description[Deep learning input pipeline with \tensorsockets using a shared CLIP model on the gpu.]{Deep learning input pipeline with \tensorsockets using a shared CLIP model on the gpu.}
\end{subfigure}
  \caption{Sharing example for generative DALL-E image generation models.}
  \label{fig:dalle_design}
\end{figure*}

In some cases, it is beneficial to move more tasks to the \textit{producer}.
For example, the training of some generative models requires pre-computed data representations in the form of embeddings for training the diffusion prior.
These embeddings are usually generated before training \cite{ramesh2022hierarchical}, \textit{offline}, but can be generated on the fly, \textit{online}, via a model inference task on the GPU. Online generation offers flexibility for new data and circumvents having to use extra disk space to store the embeddings generated a priori but is more taxing on the hardware resources while training.
\tensorsockets can move the data loading operations and the embedding generation task to its \textit{producer} as it is essentially part of the data loading pipeline. This minimizes the computational footprint on not just the CPU but also the GPU when sharing.
\Cref{fig:dalle_design} illustrates this scenario for the DALL-E model with and without sharing, and \Cref{sec:results:generative} evaluates it.

%% file: sections/4-results.tex
\section{Results}
\label{sec:results}

We now quantify the expected benefits of the data and work sharing enabled by \tensorsocket.
Our evaluation aims to answer the following questions:
\begin{list}{\labelitemi}{\leftmargin=1.5em}
\item{What is the impact of \tensorsockets on training efficiency?}
\item{What are the cost savings \tensorsockets can provide?}
\item{How do the benefits of \tensorsockets vary across different hardware setups and machine learning pipelines?}
\item{How does \tensorsockets compare to state-of-the-art data sharing solutions for model training?}
\end{list}

To answer these questions we evaluate \tensorsockets in a variety of scenarios inspired by the use cases listed in \Cref{sec:implementation:scenarios}.

\begin{table}
\centering
\begin{tabular}{ccc}
\toprule
\textbf{Application} & \textbf{Model} & \textbf{Dataset} \\ \hline
\multirow{5}{*}{\begin{tabular}[c]{@{}c@{}}Image\\ Classification\end{tabular}}
    & RegNetX 002           & \multirow{5}{*}{\begin{tabular}[c]{@{}c@{}}ImageNet\end{tabular}}          \\ 
    & RegNetX 004           &           \\ 
    & ResNet18              &           \\ 
    & MobileNetV3-Small 0.75  &           \\ 
    & MobileNetV3-Large 1.00 &           \\ \hline
\multirow{1}{*}{\begin{tabular}[c]{@{}c@{}}Audio Classification\end{tabular}}
    & CLMR          & LibriSpeech       \\ \bottomrule
\multirow{1}{*}{\begin{tabular}[c]{@{}c@{}}Image Generation\end{tabular}}
    & DALL-E 2 (Diffusion Prior)                 & CC3M       \\ \bottomrule
\multirow{1}{*}{\begin{tabular}[c]{@{}c@{}}LLM Fine-Tuning\end{tabular}}
    & Qwen2.5 0.5B                & Alpaca       \\ \bottomrule
\end{tabular}
\vspace{3mm}
\caption{Evaluated models and datasets.}
\label{tab:throughput_setup}
\end{table}

\subsection{Experimental Setup}
\label{sec:results:hardware_setup}

\textbf{Use cases}. We seek to demonstrate the value of \tensorsockets on a range of workloads that benefit from different degrees of shared data loading. We therefore evaluate DL models from the domains of computer vision, audio classification, image generation, and natural language processing. We investigate a wide range of popular computer vision models from TIMM \cite{rw2019timm}, use CLMR as our audio classification workload \cite{spijkervet2021contrastive}, source the image generation model from a well-known and tested PyTorch implementation of DALL-E 2 \cite{pw2022dalle2, ramesh2022hierarchical}, and evaluate Qwen2.5 \cite{yang2024qwen25}, a popular LLM from Alibaba, using TorchTune \cite{torchtune}.
The datasets chosen for our evaluation are ImageNet-1K \cite{deng2009imagenet}, LibriSpeech \cite{7178964}, Conceptual Captions (CC3M) \cite{sharma-etal-2018-conceptual}, and Alpaca \cite{alpaca}, respectively. \Cref{tab:throughput_setup} lists the evaluated models and the corresponding datasets.

\textbf{Hardware setup.}
We evaluate the scenarios on multiple hardware configurations.
\Cref{tab:hardware_setup} details the cloud instances and on-prem servers used in our evaluations.
The cloud configurations allow for testing the CPU utilization benefits of \tensorsockets by varying the amount of vCPUs while keeping the GPU count the same.
The A100 server features multiple GPUs allowing us to evaluate
data sharing when each GPU trains a separate model. 
Finally, the H100 server's GPU is large enough to collocate multiple DALL-E 2 training tasks.
As a result, the variety of the hardware setups allow us to evaluate the impact of \tensorsockets on different environments, use case scenarios, and collocation options.

Modern GPUs support different primitives for workload collocation on a single GPU \cite{radt}. In this work, we utilize NVIDIA Multi-Process Service (MPS) \cite{nvidia_mps_doc},
unless stated otherwise,
since it is shown to allow flexible collocation while exhibiting high performance \cite{robroek2023analysis}.
Processes executed under MPS share both GPU memory and the streaming multiprocessors (SMs). The MPS daemon automatically handles the sharing of the SMs across the collocated processes.

\begin{table}
\centering
\begin{tabular}{ccccc}
\toprule
 \textbf{Instance} & \textbf{(v)CPUs} & \textbf{GPU} & \textbf{VRAM} & \textbf{Cost} \\ \hline
 H100 Server             & 24             & H100   & 80 GB & -     \\ 
 A100 Server             & 128 (*48)             & 4x A100 & 4x 40 GB & -        \\ \bottomrule
                          AWS g5.2xlarge              & 8              & A10G  & 24 GB & \$1.212       \\ 
                          AWS g5.4xlarge              & 16             & A10G  & 24 GB & \$1.624      \\ 
                          AWS g5.8xlarge              & 32             & A10G  & 24 GB & \$2.448      \\ \bottomrule
\end{tabular}
\vspace{3mm}
\caption{On-prem servers and cloud instances used in evaluation. Costs are on demand per hour costs for corresponding cloud instances \cite{aws_cost}. The A100 server is limited to a max of 48 cores to mimic Azure offerings with A100 GPUs (\Cref{fig:vcpu_gpu_ratio}), which provide a 12:1 vCPU to GPU ratio.}
\label{tab:hardware_setup}
\end{table}

\textbf{Metrics.}
We train the same models on the same dataset without changing the learning process and thus without impacting accuracy.
Instead, we focus on the training speed and hardware utilization as the performance metrics.
We quantify the training speed via \textit{samples/s},
the amount of training samples processed by the training loop per second.
\textit{CPU Utilization}
is measured via \texttt{top} \cite{topmanual}.
We measure \textit{GPU Memory Usage} with SMI~\cite{NVIDIASystemManagement2012} and
\textit{GPU Utilization} with \texttt{dcgm}'s~\cite{dcgmmanual} SM Activity, which is shown to illustrate a finer-grained view on GPU utilization
compared to other readings from the \texttt{dcgm} tool \cite{yousefzadeh2023profiling}. Finally, we report average data movement for disk \textit{(I/O)} via \texttt{iostat}~\cite{IostatLinuxManual} and for \textit{PCIe} and \textit{NVLink} via \texttt{dcgm}.

\textbf{Training runs}.
All experiments are run with Python 3 and PyTorch 2,
and the latest versions of the respective model repositories as of writing,
using the radT platform \cite{radt}.
We ran everything twice to validate the results and
stick to the default model parameter settings as specified by the model repositories.
We set the total number of data-loading workers across the collocated workloads to the number of available CPU cores (or to 48 for the A100 server as explained in \Cref{tab:hardware_setup}).
These workers are split equally among the training processes in the experiments with conventional data loading (no sharing). Finally, we enable GPU pre-fetching for our baselines as \tensorsockets by default prefetches data to the GPU.

\begin{figure}[t]
\centering
\begin{subfigure}{.495\textwidth}
\includegraphics[width=0.95\linewidth]{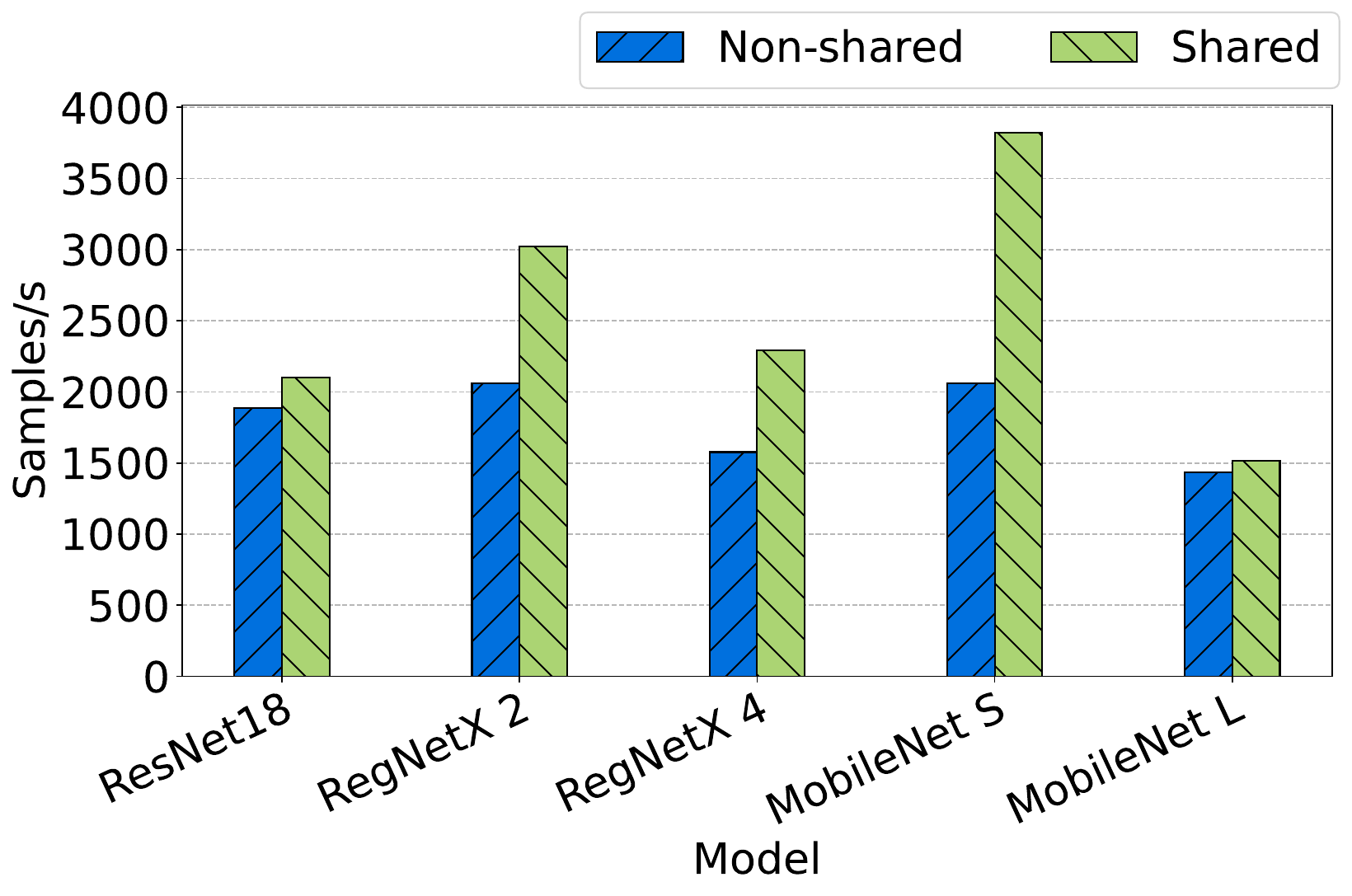}
\caption{Throughput.}
\label{fig:dgx_4way_throughput}
\end{subfigure}
\begin{subfigure}{.235\textwidth}
\includegraphics[width=1.0\linewidth]{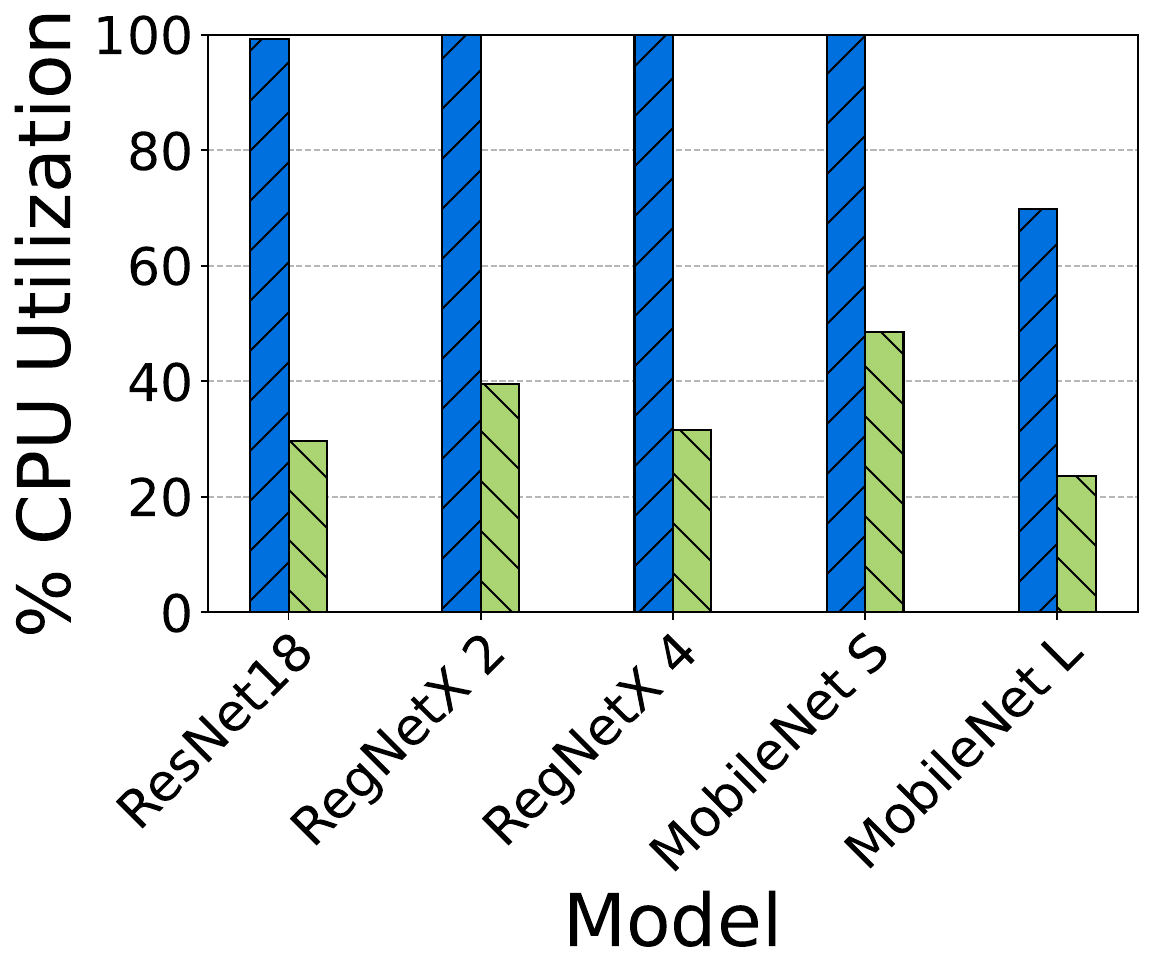}
\caption{CPU Utilization}
\label{fig:dgx_4way_CPU_util}
\end{subfigure}
\begin{subfigure}{.235\textwidth}
\includegraphics[width=1.0\linewidth]{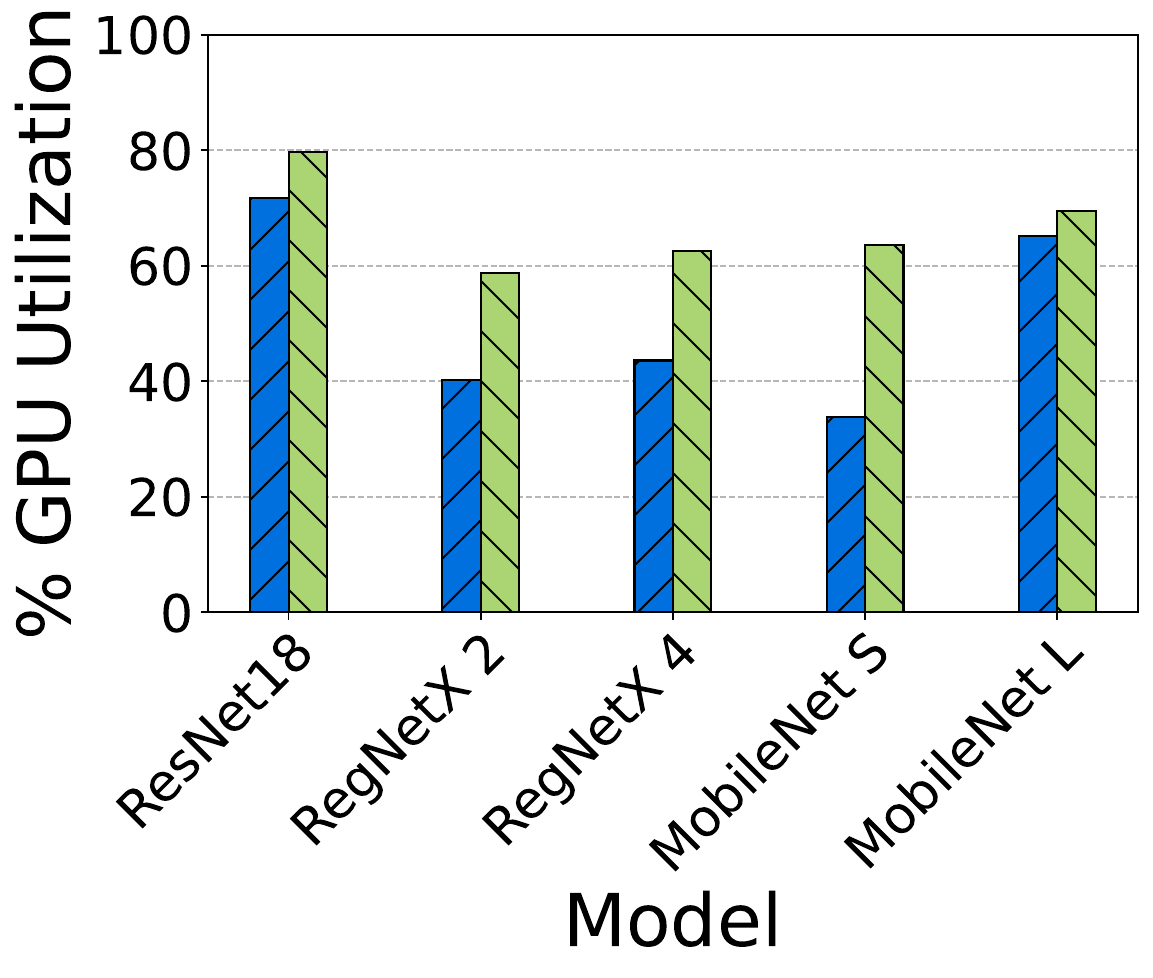}
\caption{GPU Utilization}
\label{fig:dgx_4way_GPU_util}
\end{subfigure}
\caption{Image classification training on the A100 server with 4-way collocation, where each GPU has one instance of the same model training, w/o sharing via \tensorsocket.}
\label{fig:dgx_4way}
\Description[Image classification training on the A100 server with 4-way collocation, where each GPU has one instance of the same model training, w/o sharing via \tensorsocket.]{Image classification training on the A100 server with 4-way collocation, where each GPU has one instance of the same model training, w/o sharing via \tensorsocket. \tensorsockets drastically reduces CPU requirements, maximizing throughput.}
\end{figure}

\begin{table}[t]
\begin{tabular}{l|l|l|l|l|l|}
\cline{2-6}
                                                         & \textbf{GPU} & \textbf{I/O}              & \textbf{PCIe} & \textbf{NVLink} & \textbf{VRAM} \\ \hline
\multicolumn{1}{|l|}{\multirow{4}{*}{\textbf{Baseline}}} & 0            & \multirow{4}{*}{\begin{tabular}[c]{@{}l@{}}613\\ MB/s\end{tabular}} & 270 MB/s       & -               & 8.5 GB        \\ \cline{2-2} \cline{4-6} 
\multicolumn{1}{|l|}{}                                   & 1            &                           & 267 MB/s       & -               & 8.5 GB        \\ \cline{2-2} \cline{4-6} 
\multicolumn{1}{|l|}{}                                   & 2            &                           & 268 MB/s       & -               & 8.5 GB        \\ \cline{2-2} \cline{4-6} 
\multicolumn{1}{|l|}{}                                   & 3            &                           & 267 MB/s       & -               & 8.5 GB        \\ \hline
\multicolumn{1}{|l|}{\multirow{4}{*}{\textbf{Shared}}}   & 0 (Both)     & \multirow{4}{*}{\begin{tabular}[c]{@{}l@{}}161\\ MB/s\end{tabular}} & 286 MB/s       & -               & 9.8 GB        \\ \cline{2-2} \cline{4-6} 
\multicolumn{1}{|l|}{}                                   & 1 (Cons)     &                           & 23 MB/s        & 267 MB/s        & 8.5 GB        \\ \cline{2-2} \cline{4-6} 
\multicolumn{1}{|l|}{}                                   & 2 (Cons)     &                           & 24 MB/s        & 269 MB/s        & 8.4 GB        \\ \cline{2-2} \cline{4-6} 
\multicolumn{1}{|l|}{}                                   & 3 (Cons)     &                           & 23 MB/s        & 268 MB/s        & 8.4 GB        \\ \hline
\end{tabular}
\caption{Disk I/O, CPU-to-GPU PCIe, and GPU-to-GPU NVLink traffic and GPU memory usage for four MobileNet L models training on separate A100 GPUs.}
\label{mobilenet_table}
\end{table}

\subsection{Image Classification}
\label{sec:results:image}

\hspace{\parindent}\textbf{Impact on throughput and hardware utilization.}
We first evaluate \tensorsocket's impact on the training efficiency over the most basic collocation scenario, where the same model is trained on a separate GPU available on the server
(e.g., an hyper-parameter tuning scenario).
We train a variety of image classification models, as listed in \Cref{tab:throughput_setup}, on ImageNet.
ResNet18, RegNetx~4, and MobileNet~L are more demanding models to train, while RegNetX~2 and MobileNet~S are smaller.
Among our hardware setups (\Cref{tab:hardware_setup}), the A100 server is the only one with multiple GPUs available. 
With 12 CPU-cores per GPU, this scenario additionally showcases \tensorsocket's benefits when the CPU-to-GPU ratio is too low to fully utilize the whole system, which is common among lower-price cloud offerings (\Cref{fig:vcpu_gpu_ratio}).

\Cref{fig:dgx_4way} reports the per-model training throughput and hardware utilization.
In the case of no shared data loader, the training script runs separately on each GPU.
When using \tensorsocket, we direct the producer to GPU 0 and launch a consumer on each of the four GPUs.
The producer and the consumers can communicate via NVLink, which is available on this server across the GPUs.

\tensorsockets increases the training throughput across all workloads.
In MobileNet~S's case the throughput almost doubles, whereas for models such as ResNet18 and MobileNet~L the increase ranges from $5\%$ to $10\%$.
The degree of improvement correlates with the computational complexity of the models.

For models such as ResNet18, RegNetX~2, RegNetX~4, and MobileNet~S, \Cref{fig:dgx_4way} reveals that under traditional data loading the CPU is fully utilized while the GPUs are not.
This underlies that the CPU becomes the bottleneck causing underutilization of the GPU resources.
Sharing via \tensorsockets resolves this bottleneck by reducing the stress on the CPUs while achieving a higher GPU utilization in addition to the throughput benefits. 

On the other hand, for models that are not CPU-bound such as MobileNet~L,
\tensorsockets provides marginal benefits on the throughput and GPU utilization.
However, it frees up $70\%$
of CPU resources.
The savings in CPU resources can allow for collocating additional workloads on the CPU-side in an on-prem setting and cutting the costs in a cloud setting.

\begin{figure}
\centering
\includegraphics[width=1.0\linewidth]{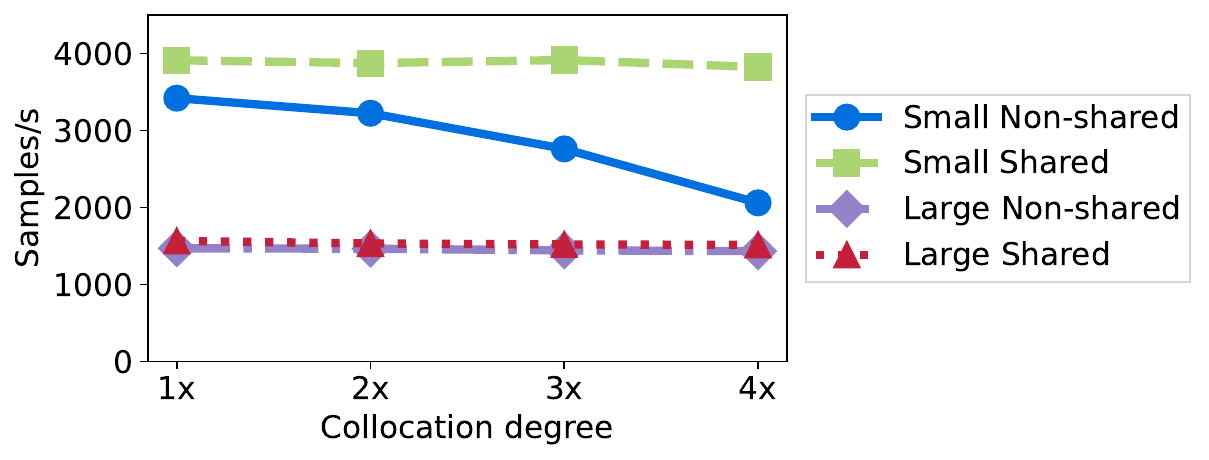}
    \caption{Per-model training throughput of MobileNet Small and Large with increasing degree of collocation on the A100 server. Each collocated model is trained on a separate GPU.}
\label{fig:dgx_4way_deepdive}
\Description[Throughput when training MobileNet with multiple degrees of collocation.]{Throughput when training MobileNet with multiple degrees of collocation. \tensorsocket's throughput does not drop when increasing collocation.}
\end{figure}

\textbf{Data movement.}
\Cref{mobilenet_table} reports the disk I/O, PCIe, and NVLink traffic in addition to GPU memory usage for 
training four MobileNet L models from \Cref{fig:dgx_4way}. 
As \tensorsocket's producer loads the data once 
and broadcasts it to all the consumers after preparing it,
it is able to greatly reduce the disk I/O and PCIe bandwidth utilization compared to the baseline, replacing it with NVLink communication instead. 
There is a small increase in VRAM usage on the producer GPU. This is due to \tensorsockets's default configuration holding some extra data ready on the GPU for e.g. buffering.

\textbf{Degree of collocation.}
We also provide a sensitivity analysis with respect to varying levels of collocation in \Cref{fig:dgx_4way_deepdive}. For this we use both MobileNets
as they are the models that exhibit the most and least benefit from \tensorsocket.
\tensorsockets yields a throughput increase for both the small and large MobileNet in all configurations.
On the other hand, increasing the amount of models trained simultaneously has little effect on the large model as the CPU is not the limiting factor (\Cref{fig:dgx_4way}).
Conversely, scaling up collocation for the small MobileNet relies on \tensorsockets to maintain high throughput.

\begin{figure}
\centering
\begin{subfigure}{.50\linewidth}
\includegraphics[width=1.0\linewidth]{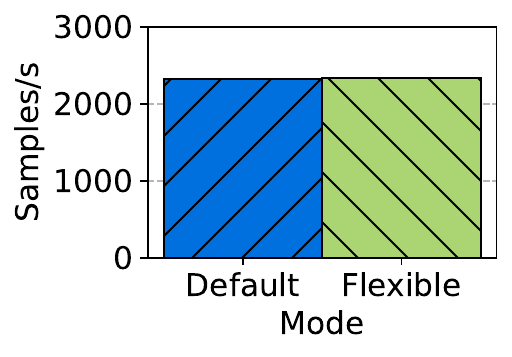}
\caption{Throughput}
\label{fig:flexible_samples}
\end{subfigure}
\begin{subfigure}{.48\linewidth}
\includegraphics[width=1.0\linewidth]{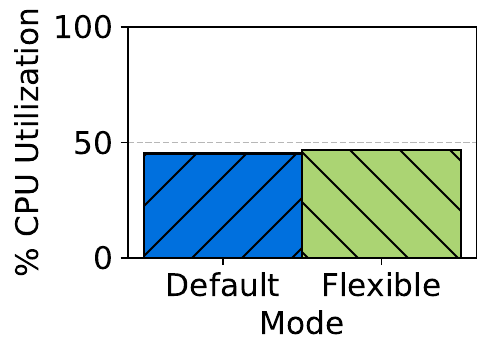}
\caption{CPU Utilization}
\label{fig:flexible_cpu}
\end{subfigure}
    \caption{Training three MobileNet Small models on the H100 server with a batch size of 128 (default) versus batch sizes of 128, 192, and 224 (flexible), following the proportions of the example distribution featured in \Cref{fig:flexible-batches}.}
\label{fig:flexible_results}
\Description[Throughput and CPU utilization of flexible batch sizing.]{Throughput and CPU utilization of flexible batch sizing. Flexible batch sizing does not significantly impact throughput and CPU utilization.}
\end{figure}

\textbf{Flexible batching.}
Finally, we evaluate the effect of flexible batches on training speed and CPU utilization. \Cref{fig:flexible_results} compares training three models with the same batch size to training three models with differing batch sizes. Flexible batching sustains training throughput while only incurring minimal CPU overhead to orchestrate the different batches required by the consumers.

\begin{figure}
\centering
\includegraphics[width=0.99\linewidth]{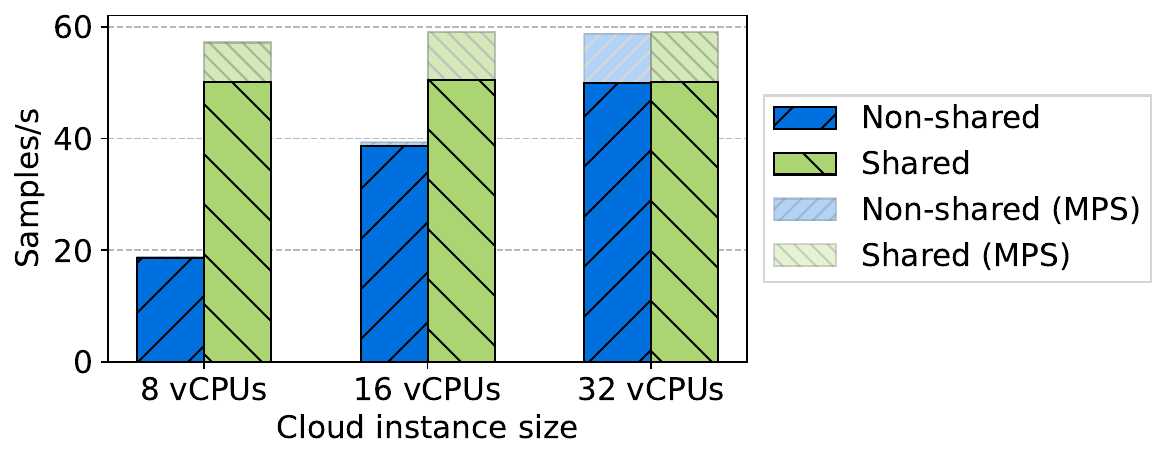}
    \caption{Samples/s per collocated training on AWS G5 Instances for CLMR - 4-way collocation on the same GPU using MPS and multi-streams across different vCPU counts w/o data sharing via \tensorsocket.}
\label{fig:g5_4way}
\Description[Samples/s per collocated training on AWS G5 Instances for CLMR.]{Samples/s per collocated training on AWS G5 Instances for CLMR. \tensorsockets allows for maximum throughput even with limited vCPU resources.}
\end{figure}

\subsection{Audio Classification}
\label{sec:results:audio}

We train the CLMR audio classification models in a 4-way collocated fashion on different AWS instances in order to showcase the impact of data and work sharing on host-side resources.
The results are shown in \Cref{fig:g5_4way}.

For this setup, in addition to MPS-based sharing, we evaluate running collocated processes as a separate GPU stream, which provides more restricted sharing, but may sometimes be the only option in a shared hardware setup such as the cloud. 
The blurred parts of the bars, therefore, highlight the additional throughput benefits of MPS-based sharing over multi-streams.
Regardless, \tensorsockets is compatible with any form of GPU sharing primitive.

From \Cref{fig:g5_4way},
on the machine with the highest number of vCPUs,
the workload with and without sharing achieve the same throughput.
This indicates that the number of vCPUs necessary to sustain the GPU throughput for this workload is met.
Without \tensorsocket, however, the smallest instance size of 8 vCPUs performs drastically worse than the largest with 32 vCPUs.
\tensorsockets
effectively reduces the amount vCPU requirement by $75\%$.
The result is that under \tensorsockets all three sizes of cloud instances achieve high training throughput.
Based on the costs reported in \Cref{tab:hardware_setup},
this leads to cloud cost savings of about $50\%$.

\subsection{Image Generation}
\label{sec:results:generative}

As mentioned in \Cref{sec:implementation:scenarios},
\tensorsockets can be used to share not just tasks on the CPU but also on the GPU.
We analyze such a pipeline using the DALL-E~2 image generation (diffusion) workload.
When training DALL-E~2, data passed to its training process must pass through a CLIP model (\Cref{sec:implementation:scenarios:generative} and \Cref{fig:dalle_design}). CLIP translates the input data into a representation that can be used by the trained model. The CLIP model can be seen as a model with frozen weights when training the diffusion model. This essentially boils down to running inference tasks on the GPU as part of the data preparation process for DALL-E training.

With \tensorsocket, we can move the CLIP model inference to the producer of the shared data loader.
In this scenario, we aim to showcase how \tensorsockets can also reduce redundancy in the computational footprint on GPUs.
By only needing a single CLIP model, we can collocate multiple DALL-E~2 diffusion models without running multiple instances of CLIP inference.
This workload is carried out on the H100 Server machine, as it is capable of supporting 4-way collocation of diffusion model training.

\begin{figure}
\centering
\includegraphics[width=0.95\linewidth]{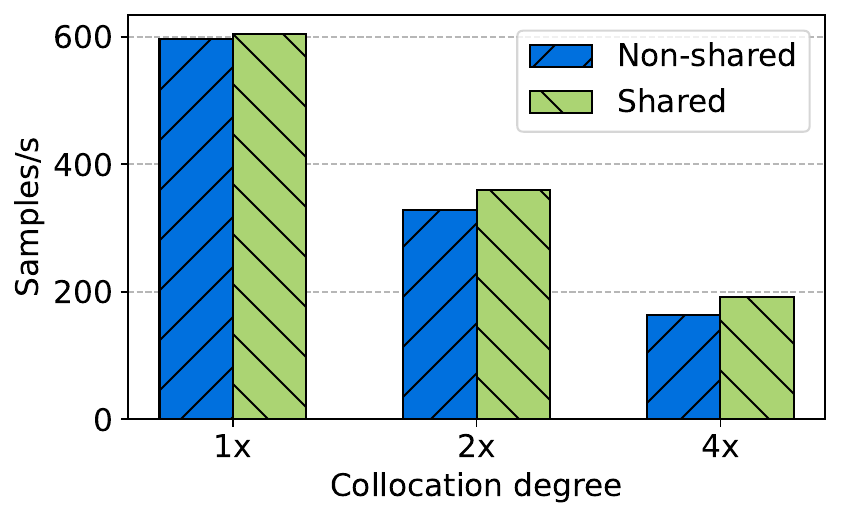}
    \caption{Samples/s per collocated online training of DALL-E on the H100 server with 1-, 2- and 4-way collocation and w/o sharing via \tensorsocket.}
\label{fig:h100_1_2_4way}
\Description[Samples/s per collocated training of DALL-E.]{Samples/s per collocated training of DALL-E. Shared training maintains more throughput under collocation.}
\end{figure}

\Cref{fig:h100_1_2_4way} shows the impact.
Since the H100 server has enough CPUs to feed the one GPU
(\Cref{tab:hardware_setup}),
this evaluation scenario is not CPU-bound.
Nevertheless,
we observe a speedup over non-shared operation under collocation.
Under 2- and 4-way collocation \tensorsockets is $10\%$ to $15\%$ faster in aggregate throughput than without when running online training.
The throughput per individual training process gets reduced
as expected,
since in this setup the GPU is highly utilized even without collocation
due to the demanding model.
This shows that \tensorsockets can enable data and work sharing not only on CPUs but also on GPUs.

\subsection{Model Selection}
\label{sec:results:mixed}

\begin{figure}
\centering
\begin{subfigure}{.2615\textwidth}
\includegraphics[width=1.0\linewidth]{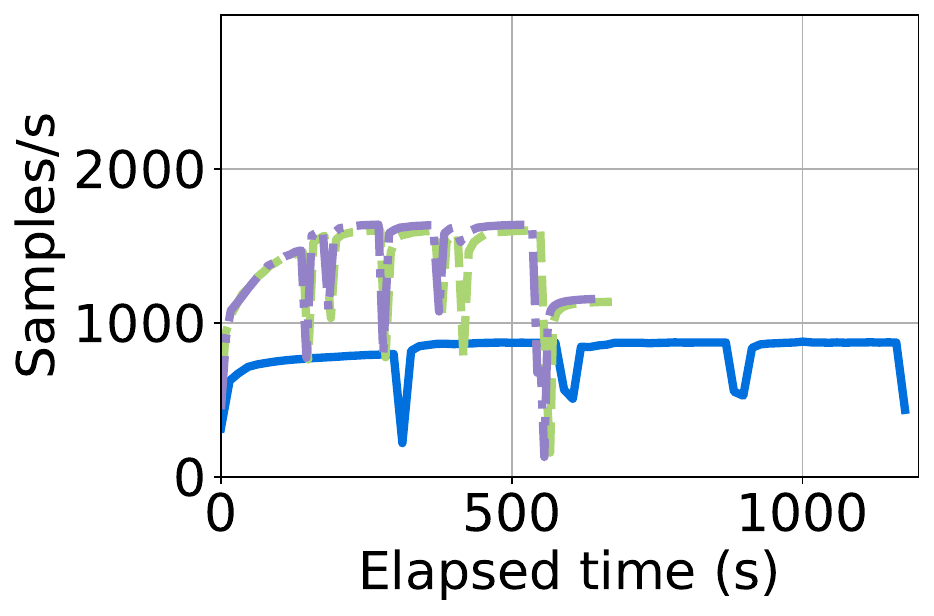}
\caption{Non-shared}
\label{fig:g5_mixed_workloads_non_shared}
\end{subfigure}
\begin{subfigure}{.2085\textwidth}
\includegraphics[width=1.0\linewidth]{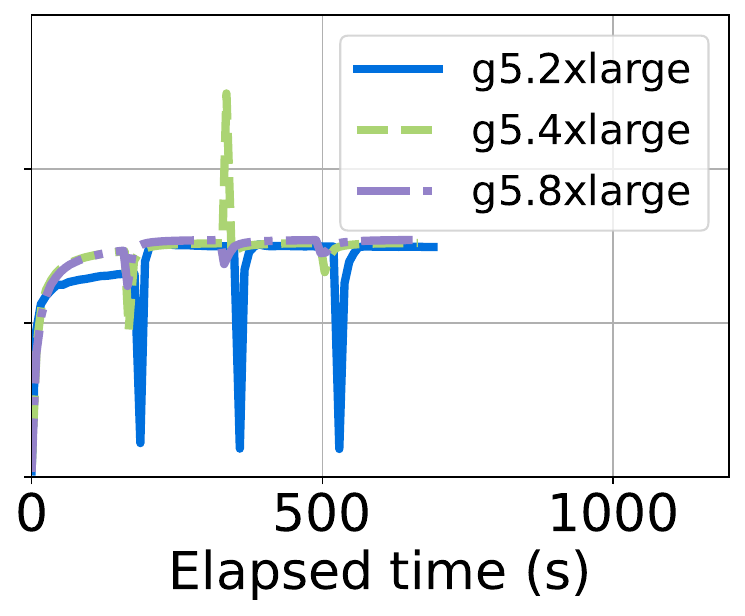}
\caption{Shared}
\label{fig:g5_mixed_workloads_shared}
\end{subfigure}
    \caption{Runtime and aggregate training throughput of mixed workloads (RegNetX~2 and RegNetX~4) on AWS G5 Instances with and without data sharing via \tensorsocket.}
\label{fig:g5_mixed_workloads}
\Description[Runtime and aggregate training throughput of mixed workloads.]{Runtime and aggregate training throughput of mixed workloads. Shared loading offers high throughput even for small instances.}
\end{figure}

In addition to the evaluations that revolved around specific domains, we show how our shared data loader supports mixed workloads, which is useful for model selection. As mentioned in \Cref{sec:implementation:scenarios:mixed}, the design of our shared data loader supports training of a mixed set of models through reallocating GPU resources between training processes. We evaluate model selection by collocating two different model training processes on different AWS cloud instances. As the training speed of the models differs, we report on the aggregate training throughput.
\Cref{fig:g5_mixed_workloads} shows the results for a mixed workload consisting of a collocated RegNetX~2 and RegNetX~4. The runs on the left use conventional non-shared data loading
whereas those on the right use \tensorsocket.
For the g5.8xlarge and g5.4xlarge AWS instances, the CPU does not constitute a bottleneck, and we therefore do not see substantial throughput gains by sharing. However, we are able to closely approximate the throughput of these larger instances with the smaller g5.2xlarge instance when sharing. In contrast, the workload throttles heavily on the small instance when not using shared data loading. For the g5.2xlarge instance, sharing is therefore not only strictly necessary for running this workload efficiently but also able to deliver almost the same throughput at half the instance cost, as seen in \Cref{tab:hardware_setup}.

\subsection{Language Model Fine-tuning}

Given the prevalence of Large Language Models (LLMs) today,
the final use case we evaluate \tensorsockets on is an LLM fine-tuning task.
Unlike prior use cases, LLM training
is generally bound on GPU compute and memory instead of data loading and CPU utilization. 
For our LLM fine-tuning task, we use the A100 server to train two Qwen2.5 models using TorchTune on the Alpaca dataset, which means \tensorsockets wraps around a Hugging Face data loader instead of a standard PyTorch one. We take the default fine-tuning recipe of TorchTune and set the batch size to 8. 
The non-sharing scenario, \textit{baseline}, runs one model per A100 GPU. The 
\tensorsockets run has the producer on a separate A100 GPU, with the two models training on GPU 1 and 2. 
This way we can observe the data traffic (PCIe and NVLink) and GPU memory use separately for the producer and the consumers.

The results are shown in \Cref{table:LLMtable}.
Due to LLMs being GPU-bound, \tensorsockets only slightly increases the training speed for this use case.
The LLMs being trained require a low amount of PCIe communication (48 MB/s on average) even for the baseline.
\tensorsocket's producer reveals
that only 341 KB/s is required to run the data loader. 
The NVLink utilization, which is used by \tensorsockets to directly communicate training data between GPUs, is only $\sim$150KB/s.
This data loading is insignificant compared to the rest of the PCIe traffic.
Finally, the GPU memory utilization for the models training on GPUs 1 and 2 is the same for the baseline and the shared case, indicating that there is no GPU memory overhead for the \tensorsockets consumers. There is a small 1.5 GB memory requirement for running the producer on GPU 0. This is a combination of CUDA overhead and \tensorsockets with the default configuration of buffering data, similar to results in \Cref{mobilenet_table}.

\begin{table}
\begin{tabular}{l|l|l|l|l|l|}
\cline{2-6}
                                                         & \textbf{GPU} & \textbf{Tokens} & \textbf{PCIe} & \textbf{NVLink} & \textbf{VRAM} \\ \hline
\multicolumn{1}{|l|}{\multirow{2}{*}{\textbf{Baseline}}} & 1            & 7.5k/s              & 48 MB/s        & -               & 7.3 GB        \\ \cline{2-6} 
\multicolumn{1}{|l|}{}                                   & 2            & 7.4k/s              & 48 MB/s        & -               & 7.3 GB        \\ \hline
\multicolumn{1}{|l|}{\multirow{3}{*}{\textbf{Shared}}}   & 0 (Prod)     & -                 & .3 MB/s        & -               & 1.5 GB        \\ \cline{2-6} 
\multicolumn{1}{|l|}{}                                   & 1 (Cons)     & 7.5k/s              & 48 MB/s        & 152 KB/s        & 7.3 GB        \\ \cline{2-6} 
\multicolumn{1}{|l|}{}                                   & 2 (Cons)     & 7.6k/s             & 48 MB/s        & 153 KB/s        & 7.3 GB        \\ \hline
\end{tabular}
\caption{Training speed, PCIe and NVLink traffic, and GPU memory usage for Qwen2.5 0.5B LLM training using TorchTune on separate A100 GPUs.}
\label{table:LLMtable}
\end{table}

\subsection{Comparison to other sharing techniques}
\label{sec:results:comp}

After having assessed the value of shared data loading via \tensorsockets in deep learning training, 
we compare \tensorsockets to state-of-the-art methods,
CoorDL \cite{mohan2021analyzing} and Joader \cite{xuDeepLearningDataloader2022},
that achieve data loading speedups via sharing. 
\footnote{While we point out the many difficulties in establishing a fair comparison across all the codebases in the rest of this section, we are grateful for the authors of both CoorDL \cite{mohan2021analyzing} and Joader \cite{xuDeepLearningDataloader2022} for providing an open-source implementation.}

Comparing to CoorDL surfaces a couple of challenges due to the age of the library \cite{coordlcodebase}.
CoorDL has been designed as a plugin for NVIDIA DALI and is written for Python 3.6. 
This version of Python, however, is deprecated by PyTorch since 2021, which means modern versions of the framework are not compatible with it. \tensorsockets is incompatible with PyTorch 1 as the deep learning framework made sweeping changes with the introduction of PyTorch 2. This complicates establishing a fair comparison between CoorDL and \tensorsocket.
Nevertheless,
we run CoorDL using the evaluation script provided by the authors of the original work \cite{mohan2021analyzing} to run it as efficiently as possible.
We adjust the parameters for \tensorsockets accordingly.
This means that automatic mixed precision is disabled, the batch size is set to 512, and there are 4 data loading workers.
We also choose ResNet18 as the evaluated model following CoorDL's evaluation.
Finally, while reporting the results, we normalize the per-model training throughput and hardware utilization values by dividing them by the values achieved by single model training (no-collocation).
This normalization is to further eliminate the impact of any unfair differences between the diverging libraries of the corresponding codebases.

\begin{figure}[t]
\centering
\begin{subfigure}{.405\textwidth}
\includegraphics[width=1.0\linewidth]{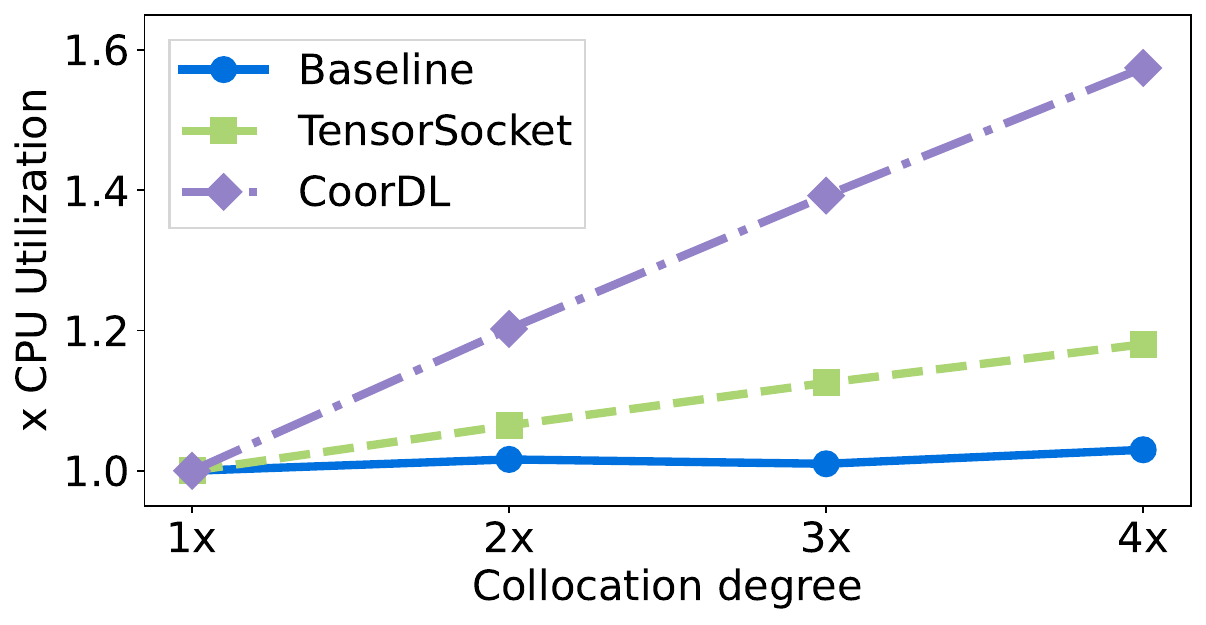}
\caption{Normalized CPU utilization.}
\label{fig:coordl_cpu}
\end{subfigure}
\begin{subfigure}{.405\textwidth}
\includegraphics[width=1.0\linewidth]{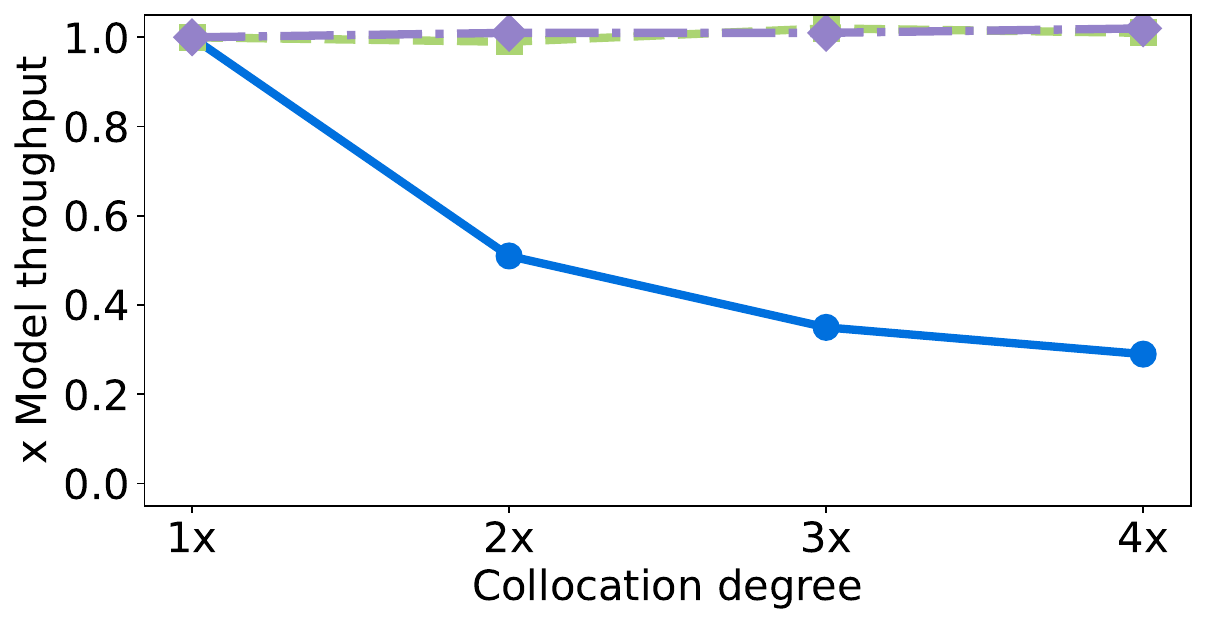}
\caption{Normalized per-model training throughput.}
\label{fig:coordl_throughput}
\end{subfigure}
    \caption{CPU utilization and throughput scaling of baseline (no-sharing), CoorDL, and \tensorsockets on the A100 system under different levels of collocation. The scaling is compared to training a single model with that technique without collocation. Each collocated ResNet18 model runs on a separate GPU. \tensorsockets is able to achieve the maximum throughput without the deployment restrictions and higher CPU requirement of CoorDL.}
\label{fig:coordl}
\Description[Scaling of CoorDL versus \tensorsocket.]{Scaling of CoorDL versus \tensorsocket. \tensorsockets matches CoorDL's performance with a fraction of the CPU cost.}
\end{figure}

\Cref{fig:coordl} reports the result of the comparison.
These experiments utilize the A100 machine with 4 GPUs.
Each instance of the ResNet18 model is being trained on ImageNet and on a separate GPU
(as in \Cref{sec:results:image}).
\Cref{fig:coordl_cpu} notes the scaling of CPU utilization as collocation increases. \tensorsockets only marginally increases CPU load under higher degrees of collocation, while CoorDL requires more CPU resources to keep up. The CPU utilization of our baseline, not using either CoorDL or \tensorsocket, is close to constant. This can be explained by \Cref{fig:coordl_throughput}, which shows the throughput scaling as the degree of collocation increases. Both CoorDL and \tensorsocket have no issue keeping the per-model training throughput the same despite the higher load. The baseline, however, heavily throttles, losing almost $75\%$ of the performance under 4x load. This throttling can explain the low CPU utilization, as the data loading workers can not keep up with the training loops and thus the models idle.
In general, while both \tensorsockets and CoorDL can provide maximum throughput, \tensorsockets does so with considerably fewer CPU resources while also being more flexible and less intrusive in usage.

\begin{figure*}[ht!]
\centering
\includegraphics[width=0.99\linewidth]{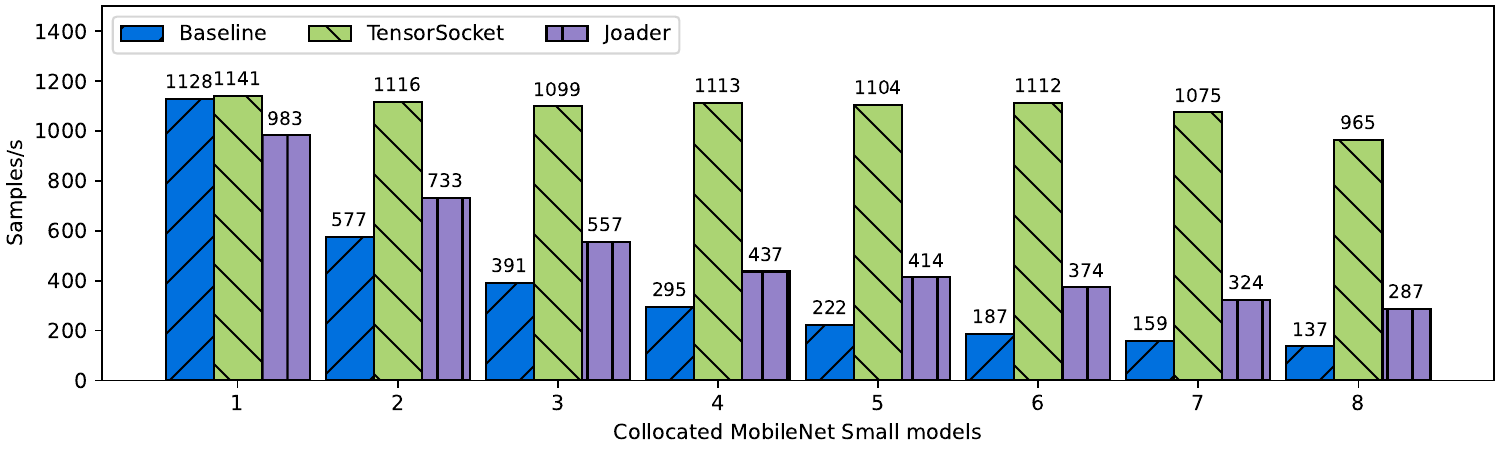}
    \caption{Comparison of model training throughput with varying degrees of collocation under constrained CPU resources on the H100 system between baseline (no data sharing), Joader \cite{xuDeepLearningDataloader2022}, and \tensorsocket. Each collocated model training is MobileNetV3-Small trained on ImageNet. \tensorsockets is able to provide data to many collocated models with little impact on throughput, even under constrained CPU circumstances.}
\label{fig:joader}
\Description[Scaling of Joader versus \tensorsocket.]{Scaling of Joader versus \tensorsocket. \tensorsockets maximizes throughput regardless of the collocation, outperforming both the baseline and Joader.}
\end{figure*}

Joader has a proof-of-concept implementation in Rust that is compatible with the latest version of PyTorch \cite{joadercodebase}. Specifically, the current implementation of Joader does not require PyTorch at all; rather, it uses Numpy to store data instead. While this does improve the compatibility of the implementation, it raises serious performance issues that hamper its effectiveness in real-world scenarios. Specifically, 1) Joader's image pre-processing and dataset (ImageNet) support is fully hardcoded, 2) images are delivered as NumPy matrices instead of Tensors and 3) Joader does not have support for mini-batches.
Therefore, we take a number of concessions in order to support a comparison to Joader that is as fair as possible. Firstly, for (1), we investigate Joader's pre-processing pipeline and configure our TIMM training script to use the same transformations that are hardcoded for Joader. Note that it is not possible to use the exact same transformation code in both, as Joader is written in Rust and pre-processing pipelines available in online model sources like TIMM are typically defined in Python. Then, we address (2) and (3) by having the training script in Joader's case ask for enough data to fill up the batch after which it can construct the tensor and send it to the GPU. This tensor construction from many NumPy matrices, however, is very expensive and will cripple the performance. As for the sake of this performance comparison we are not interested in the loss of the model, we opt to only construct a tensor out of the first batch of data. Subsequent iterations then wait for the new data to come in, only to train on the same, first batch again. This minimizes the overhead of batching for Joader.

\Cref{fig:joader} shows the results of comparing Joader this way to \tensorsockets on the H100 system. We maximize the performance of all techniques by using MPS for sharing the resources of the H100 GPU. As the baseline, we train 1 to 8 collocated MobileNetV3's without any sharing enabled. Furthermore, we restrict the amount of data loading workers to 8. This means that, under no sharing, every model only has 1 worker when training 8 models simultaneously. Collocation amounts that do not divide 8, such as 5, have the workers divided unevenly in order to sum to 8. Keep in mind that the training scripts themselves are allowed to use further CPU resources.
As expected, \tensorsockets performs well even when running with high amounts of collocation, dropping no performance up to and including 6-way collocation. Only with 7- and 8-way collocation is there a drop in performance. This comes in stark contrast to non-shared training, where throughput goes down quickly. In fact, the data loading is such a bottleneck for non-shared training here that the summed throughput of collocation never exceeds that of non-collocated training. Joader comfortably outperforms non-shared training but is far behind the efficiency of \tensorsocket. This is likely caused by the extra overhead introduced by Joader's data sampling algorithm (as described in \Cref{sec:background}). While Joader's algorithm provides flexibility for training with different speeds and datasets, 
this sacrifices efficiency in terms of CPU resource use, 
which in turn impacts the efficiency of training itself.

%% file: sections/5-discussion.tex
\section{\tensorsockets Going Forward}
\label{sec:vision}

Having evaluated \tensorsocket's impact on training efficiency, cost savings, and flexibility, 
this section highlights the key results while discussing further applicability of \tensorsocket.

\textbf{Target Workloads.}
\tensorsockets alleviates a range of computational and resource-dependent bottlenecks.
In general, small scale DL training workloads that are prone to exhibit input-bound pipelines or a high degree of CPU utilization can benefit greatly from shared data loading. If used on workloads in the cloud with similar characteristics as those presented in our work,
which are prevalent in practice \cite{hypesustainability},
\tensorsockets may reduce costs up to a substantial 50\% reduction as shown in \Cref{sec:results}. 

Achieving these benefits does require some conditions to be met. Training jobs have to be collocated to make use of a shared data loader and are required to train with similar speed on the same dataset. Loosening up these requirements may allow shared data loading to become an attractive option for more workloads, though, that may come at the cost of reduced benefits.

\textbf{Generalizability.}
We analyzed a complex data loading pipeline that includes generating intermediate representations of our data through an auxiliary model that is not being trained in \Cref{sec:results:generative}.
The ability to support unusual steps like these is a testament to the generalizability of \tensorsocket. 
In addition, we can further dissect the data loading pipeline for finer-grained sharing.
This allows for transformations and augmentations that are specific to each training process while only doing costly work, such as image decoding, once.
For other techniques that use GPUs for data pre-processing \cite{nvidia_dali, fusionflow, fastflow}, this means that \tensorsockets can be deployed together with them to support GPU-offloading of transformation and augmentation operations while keeping redundancy and computational footprint low.

\tensorsockets is implemented around PyTorch 2 and has no other large dependencies.
We leave as much of the implementation as possible, such as the data structures, over to PyTorch to minimize the complexity of our codebase. This makes \tensorsockets easily maintainable for future PyTorch versions and extendable. Other frameworks, such as TensorFlow, can use \tensorsockets now by using PyTorch as a data sharing intermediate.
A native implementation in other frameworks 
would require just the re-implementation of the wrapper that allows for tensor deconstruction and reconstruction, \textit{TensorPayload}, estimated $\sim$59 lines of code.

%% file: sections/6-related-work.tex
\section{Related Work}
\label{sec:related_work}

Historically, there has been a plethora of work on benchmarking and optimizing the computational efficiency of the core of model training and serving.
However, in the recent years, 
data loading and pre-processing have been 
gaining more attention \cite{Balmau22, mlcommonsstorage},
as with ever-increasing model sizes and training throughput requirements the cost of data loading bottlenecks is growing rapidly \cite{metaDataLoadingAnalysis}.

Murray et al. \cite{tf.data} and Mohan et al. \cite{mohan2021analyzing}
emphasize the impact of the data loading pipelines on training efficiency.
The former presents the tf.data framework to ease the tuning for the computational efficiency of data pre-processing.
The latter proposes CoorDL, a data loading library, and MinIO, a software cache with the goal of reducing cache thrashing.
These works, among others \cite{234827, audibert2023tf}, also advocate for sharing for DL data loading tasks, but more at the cluster-level rather than the finer-grained view of \tensorsocket.

To further optimize data loading flows, Plumber \cite{kuchnik_plumber_2022} extends tf.data by automatically detecting and resolving misconfigured pipelines, Lotus and Presto provides data pipeline optimization insights \cite{bachkaniwalaLotusCharacterizationMachine2024, isenko}, while Cedar \cite{zhao_cedar_2024} and Pecan \cite{graurPecanCostEfficientML} orchestrate optimizations to the input data pipeline.
Joader, proposed by Xu et al. \cite{xuDeepLearningDataloader2022}, provides an alternative to CoorDL that offers extra flexibility when sharing data for training tasks, allowing for shared training on multiple datasets at the same time. It manages this via a novel data sampling solution that optimizes sharing at the cost of repeated dataset intersection calculations.
Such calculations can be costly, as \Cref{sec:results:comp} demonstrated.

Behme et al. \cite{behme_art_2023} explore lossy image compression's role in mitigating data loading bottlenecks. They demonstrate that moderately compressed data maintains accuracy comparable to benchmarks while saving 30\% storage and how this technique complements the software cache of MinIO \cite{mohan2021analyzing}.

\tensorsockets also borrows ideas from works on data and work sharing in databases such as StagedDB \cite{stageddb}, QPipe \cite{qpipe}, and SharedDB \cite{shareddb}, and works that analyze the trade-offs of sharing \cite{psaroudakisAA13, JohnsonHPMHSAF07}.
These works aim at sharing work that is common across concurrent database queries.
In contrast, \tensorsockets applies similar sharing ideas to DL data preparation.

%% file: sections/7-conclusion.tex
\section{Conclusion}
\label{sec:conclusion}

In this paper,
we presented \tensorsocket,
a novel data loading mechanism that enables data and work sharing in data pre-processing pipelines of deep learning training.
The key insight behind \tensorsockets is that tuning efforts to achieve the best model architecture and parameters require training several models on the same data.
This results in shared tasks across these training processes.
We demonstrated that
\tensorsockets
can double training throughput while substantially reducing the number of CPU cores to achieve that throughput.
Furthermore, it
is easy to adopt in existing training pipelines,
enables certain training scenarios on restricted hardware resource setups,
and can halve cloud setup costs as a result.
Finally, 
\tensorsockets is compatible with existing techniques that aim at increasing the computational efficiency of the data pre-processing tasks.

%% file: main.bbl

\begin{thebibliography}{72}


\ifx \showCODEN    \undefined \def \showCODEN     #1{\unskip}     \fi
\ifx \showDOI      \undefined \def \showDOI       #1{#1}\fi
\ifx \showISBNx    \undefined \def \showISBNx     #1{\unskip}     \fi
\ifx \showISBNxiii \undefined \def \showISBNxiii  #1{\unskip}     \fi
\ifx \showISSN     \undefined \def \showISSN      #1{\unskip}     \fi
\ifx \showLCCN     \undefined \def \showLCCN      #1{\unskip}     \fi
\ifx \shownote     \undefined \def \shownote      #1{#1}          \fi
\ifx \showarticletitle \undefined \def \showarticletitle #1{#1}   \fi
\ifx \showURL      \undefined \def \showURL       {\relax}        \fi
\providecommand\bibfield[2]{#2}
\providecommand\bibinfo[2]{#2}
\providecommand\natexlab[1]{#1}
\providecommand\showeprint[2][]{arXiv:#2}

\bibitem[{Amazon}(2023a)]%
        {aws_ratio}
\bibfield{author}{\bibinfo{person}{{Amazon}}.} \bibinfo{year}{2023}\natexlab{a}.
\newblock \bibinfo{howpublished}{\url{https://aws.amazon.com/ec2/}}.
\newblock
\newblock
\shownote{Accessed: 2023-11-30}.


\bibitem[{Amazon}(2023b)]%
        {aws_cost}
\bibfield{author}{\bibinfo{person}{{Amazon}}.} \bibinfo{year}{2023}\natexlab{b}.
\newblock \bibinfo{howpublished}{\url{https://aws.amazon.com/ec2/pricing/on-demand/}}.
\newblock
\newblock
\shownote{Accessed: 2024-01-22}.


\bibitem[Anthony et~al\mbox{.}(2020)]%
        {carbontracker}
\bibfield{author}{\bibinfo{person}{Lasse F.~Wolff Anthony}, \bibinfo{person}{Benjamin Kanding}, {and} \bibinfo{person}{Raghavendra Selvan}.} \bibinfo{year}{2020}\natexlab{}.
\newblock \bibinfo{title}{Carbontracker: Tracking and Predicting the Carbon Footprint of Training Deep Learning Models}.
\newblock \bibinfo{howpublished}{ICML Workshop on Challenges in Deploying and monitoring Machine Learning Systems}.
\newblock
\newblock
\shownote{arXiv:2007.03051}.


\bibitem[Audibert et~al\mbox{.}(2023)]%
        {audibert2023tf}
\bibfield{author}{\bibinfo{person}{Andrew Audibert}, \bibinfo{person}{Yang Chen}, \bibinfo{person}{Dan Graur}, \bibinfo{person}{Ana Klimovic}, \bibinfo{person}{Ji{\v{r}}{\'\i} {\v{S}}im{\v{s}}a}, {and} \bibinfo{person}{Chandramohan~A Thekkath}.} \bibinfo{year}{2023}\natexlab{}.
\newblock \showarticletitle{tf.data service: A Case for Disaggregating ML Input Data Processing}. In \bibinfo{booktitle}{\emph{Proceedings of the 2023 ACM Symposium on Cloud Computing}}. \bibinfo{publisher}{Association for Computing Machinery}, \bibinfo{address}{USA}, \bibinfo{pages}{358--375}.
\newblock


\bibitem[Bachkaniwala et~al\mbox{.}(2024)]%
        {bachkaniwalaLotusCharacterizationMachine2024}
\bibfield{author}{\bibinfo{person}{Rajveer Bachkaniwala}, \bibinfo{person}{Harshith Lanka}, \bibinfo{person}{Kexin Rong}, {and} \bibinfo{person}{Ada Gavrilovska}.} \bibinfo{year}{2024}\natexlab{}.
\newblock \showarticletitle{Lotus: {Characterization} of {Machine} {Learning} {Preprocessing} {Pipelines} via {Framework} and {Hardware} {Profiling}}. In \bibinfo{booktitle}{\emph{2024 {IEEE} {International} {Symposium} on {Workload} {Characterization} ({IISWC})}}. \bibinfo{publisher}{IEEE}, \bibinfo{address}{Vancouver, BC, Canada}, \bibinfo{pages}{30--43}.
\newblock
\showISBNx{979-8-3503-5603-8}
\urldef\tempurl%
\url{https://doi.org/10.1109/IISWC63097.2024.00013}
\showDOI{\tempurl}


\bibitem[Balmau(2022)]%
        {Balmau22}
\bibfield{author}{\bibinfo{person}{Oana Balmau}.} \bibinfo{year}{2022}\natexlab{}.
\newblock \showarticletitle{Characterizing {I/O} in Machine Learning with MLPerf Storage}.
\newblock \bibinfo{journal}{\emph{{SIGMOD} Rec.}} \bibinfo{volume}{51}, \bibinfo{number}{3} (\bibinfo{year}{2022}), \bibinfo{pages}{47--48}.
\newblock
\urldef\tempurl%
\url{https://doi.org/10.1145/3572751.3572765}
\showDOI{\tempurl}


\bibitem[Baunsgaard et~al\mbox{.}(2020)]%
        {baunsgaardWT20}
\bibfield{author}{\bibinfo{person}{Sebastian Baunsgaard}, \bibinfo{person}{Sebastian~Benjamin Wrede}, {and} \bibinfo{person}{Pınar Tözün}.} \bibinfo{year}{2020}\natexlab{}.
\newblock \showarticletitle{{Training for Speech Recognition on Coprocessors}}. In \bibinfo{booktitle}{\emph{International Workshop on Accelerating Analytics and Data Management Systems Using Modern Processor and Storage Architectures, ADMS@VLDB 2020, Tokyo, Japan, August 31, 2020}}, \bibfield{editor}{\bibinfo{person}{Rajesh Bordawekar} {and} \bibinfo{person}{Tirthankar Lahiri}} (Eds.). \bibinfo{publisher}{Association for Computing Machinery}, \bibinfo{address}{Japan}, \bibinfo{pages}{1--10}.
\newblock


\bibitem[Behme et~al\mbox{.}(2023)]%
        {behme_art_2023}
\bibfield{author}{\bibinfo{person}{Lennart Behme}, \bibinfo{person}{Saravanan Thirumuruganathan}, \bibinfo{person}{Alireza~Rezaei Mahdiraji}, \bibinfo{person}{Jorge{-}Arnulfo Quian\'{e}{-}Ruiz}, {and} \bibinfo{person}{Volker Markl}.} \bibinfo{year}{2023}\natexlab{}.
\newblock \showarticletitle{The Art of Losing to Win: Using Lossy Image Compression to Improve Data Loading in Deep Learning Pipelines}. In \bibinfo{booktitle}{\emph{Proceedings of the 39th IEEE International Conference on Data Engineering}}. \bibinfo{publisher}{IEEE}, \bibinfo{address}{Anaheim, California}, \bibinfo{pages}{936--949}.
\newblock


\bibitem[Cubuk et~al\mbox{.}(2019)]%
        {cubuk2019autoaugment}
\bibfield{author}{\bibinfo{person}{Ekin~D. Cubuk}, \bibinfo{person}{Barret Zoph}, \bibinfo{person}{Dandelion Mane}, \bibinfo{person}{Vijay Vasudevan}, {and} \bibinfo{person}{Quoc~V. Le}.} \bibinfo{year}{2019}\natexlab{}.
\newblock \showarticletitle{AutoAugment: Learning Augmentation Strategies From Data}. In \bibinfo{booktitle}{\emph{Proceedings of the IEEE/CVF Conference on Computer Vision and Pattern Recognition (CVPR)}}. \bibinfo{publisher}{IEEE}, \bibinfo{address}{USA}, \bibinfo{pages}{113--123}.
\newblock


\bibitem[Delimitrou and Kozyrakis(2014)]%
        {DelimitrouK14}
\bibfield{author}{\bibinfo{person}{Christina Delimitrou} {and} \bibinfo{person}{Christos Kozyrakis}.} \bibinfo{year}{2014}\natexlab{}.
\newblock \showarticletitle{{Quasar: Resource-Efficient and QoS-Aware Cluster Management}}. In \bibinfo{booktitle}{\emph{ASPLOS}}. \bibinfo{publisher}{Association for Computing Machinery}, \bibinfo{address}{USA}, \bibinfo{pages}{127--144}.
\newblock


\bibitem[Deng et~al\mbox{.}(2009)]%
        {deng2009imagenet}
\bibfield{author}{\bibinfo{person}{Jia Deng}, \bibinfo{person}{Wei Dong}, \bibinfo{person}{Richard Socher}, \bibinfo{person}{Li-Jia Li}, \bibinfo{person}{Kai Li}, {and} \bibinfo{person}{Li Fei-Fei}.} \bibinfo{year}{2009}\natexlab{}.
\newblock \showarticletitle{Imagenet: A large-scale hierarchical image database}. In \bibinfo{booktitle}{\emph{2009 IEEE conference on computer vision and pattern recognition}}. Ieee, \bibinfo{publisher}{IEEE}, \bibinfo{address}{USA}, \bibinfo{pages}{248--255}.
\newblock


\bibitem[Dodge et~al\mbox{.}(2022)]%
        {carbonIntensity}
\bibfield{author}{\bibinfo{person}{Jesse Dodge}, \bibinfo{person}{Taylor Prewitt}, \bibinfo{person}{Remi Tachet~des Combes}, \bibinfo{person}{Erika Odmark}, \bibinfo{person}{Roy Schwartz}, \bibinfo{person}{Emma Strubell}, \bibinfo{person}{Alexandra~Sasha Luccioni}, \bibinfo{person}{Noah~A. Smith}, \bibinfo{person}{Nicole DeCario}, {and} \bibinfo{person}{Will Buchanan}.} \bibinfo{year}{2022}\natexlab{}.
\newblock \showarticletitle{Measuring the Carbon Intensity of AI in Cloud Instances}. In \bibinfo{booktitle}{\emph{Proceedings of the 2022 ACM Conference on Fairness, Accountability, and Transparency}} (Seoul, Republic of Korea) \emph{(\bibinfo{series}{FAccT '22})}. \bibinfo{publisher}{Association for Computing Machinery}, \bibinfo{address}{New York, NY, USA}, \bibinfo{pages}{1877–1894}.
\newblock
\showISBNx{9781450393522}
\urldef\tempurl%
\url{https://doi.org/10.1145/3531146.3533234}
\showDOI{\tempurl}


\bibitem[Espenshade et~al\mbox{.}(2024)]%
        {10.1145/3642970.3655830}
\bibfield{author}{\bibinfo{person}{Connor Espenshade}, \bibinfo{person}{Rachel Peng}, \bibinfo{person}{Eumin Hong}, \bibinfo{person}{Max Calman}, \bibinfo{person}{Yue Zhu}, \bibinfo{person}{Pritish Parida}, \bibinfo{person}{Eun~Kyung Lee}, {and} \bibinfo{person}{Martha~A. Kim}.} \bibinfo{year}{2024}\natexlab{}.
\newblock \showarticletitle{Characterizing Training Performance and Energy for Foundation Models and Image Classifiers on Multi-Instance GPUs}. In \bibinfo{booktitle}{\emph{Proceedings of the 4th Workshop on Machine Learning and Systems}} (Athens, Greece) \emph{(\bibinfo{series}{EuroMLSys '24})}. \bibinfo{publisher}{Association for Computing Machinery}, \bibinfo{address}{New York, NY, USA}, \bibinfo{pages}{47–55}.
\newblock
\showISBNx{9798400705410}
\urldef\tempurl%
\url{https://doi.org/10.1145/3642970.3655830}
\showDOI{\tempurl}


\bibitem[Giannikis et~al\mbox{.}(2012)]%
        {shareddb}
\bibfield{author}{\bibinfo{person}{Georgios Giannikis}, \bibinfo{person}{Gustavo Alonso}, {and} \bibinfo{person}{Donald Kossmann}.} \bibinfo{year}{2012}\natexlab{}.
\newblock \showarticletitle{SharedDB: Killing One Thousand Queries with One Stone}.
\newblock \bibinfo{journal}{\emph{Proc. VLDB Endow.}} \bibinfo{volume}{5}, \bibinfo{number}{6} (\bibinfo{date}{feb} \bibinfo{year}{2012}), \bibinfo{pages}{526–537}.
\newblock
\showISSN{2150-8097}
\urldef\tempurl%
\url{https://doi.org/10.14778/2168651.2168654}
\showDOI{\tempurl}


\bibitem[{Google}(2023)]%
        {gcp_ratio}
\bibfield{author}{\bibinfo{person}{{Google}}.} \bibinfo{year}{2023}\natexlab{}.
\newblock \bibinfo{howpublished}{\url{https://cloud.google.com/compute?hl=en}}.
\newblock
\newblock
\shownote{Accessed: 2023-11-30}.


\bibitem[Graur et~al\mbox{.}(2022)]%
        {cachew}
\bibfield{author}{\bibinfo{person}{Dan Graur}, \bibinfo{person}{Damien Aymon}, \bibinfo{person}{Dan Kluser}, \bibinfo{person}{Tanguy Albrici}, \bibinfo{person}{Chandramohan~A. Thekkath}, {and} \bibinfo{person}{Ana Klimovic}.} \bibinfo{year}{2022}\natexlab{}.
\newblock \showarticletitle{Cachew: Machine Learning Input Data Processing as a Service}. In \bibinfo{booktitle}{\emph{2022 USENIX Annual Technical Conference (USENIX ATC 22)}}. \bibinfo{publisher}{USENIX Association}, \bibinfo{address}{Carlsbad, CA}, \bibinfo{pages}{689--706}.
\newblock
\showISBNx{978-1-939133-29-43}
\urldef\tempurl%
\url{https://www.usenix.org/conference/atc22/presentation/graur}
\showURL{%
\tempurl}


\bibitem[Graur et~al\mbox{.}(2024)]%
        {graurPecanCostEfficientML}
\bibfield{author}{\bibinfo{person}{Dan Graur}, \bibinfo{person}{Oto Mraz}, \bibinfo{person}{Muyu Li}, \bibinfo{person}{Mohammad~Sepehr Pourghannad}, \bibinfo{person}{Chandramohan~A. Thekkath}, {and} \bibinfo{person}{Ana Klimovic}.} \bibinfo{year}{2024}\natexlab{}.
\newblock \showarticletitle{Pecan: Cost-Efficient {ML} Data Preprocessing with Automatic Transformation Ordering and Hybrid Placement}. In \bibinfo{booktitle}{\emph{Proceedings of the 2024 {USENIX} Annual Technical Conference, {USENIX} {ATC} 2024, Santa Clara, CA, USA, July 10-12, 2024}}, \bibfield{editor}{\bibinfo{person}{Saurabh Bagchi} {and} \bibinfo{person}{Yiying Zhang}} (Eds.). \bibinfo{publisher}{{USENIX} Association}, \bibinfo{address}{USA}, \bibinfo{pages}{649--665}.
\newblock
\urldef\tempurl%
\url{https://www.usenix.org/conference/atc24/presentation/graur}
\showURL{%
\tempurl}


\bibitem[Harizopoulos and Ailamaki(2005)]%
        {stageddb}
\bibfield{author}{\bibinfo{person}{Stavros Harizopoulos} {and} \bibinfo{person}{Anastassia Ailamaki}.} \bibinfo{year}{2005}\natexlab{}.
\newblock \showarticletitle{StagedDB: Designing Database Servers for Modern Hardware}.
\newblock \bibinfo{journal}{\emph{{IEEE} Data Eng. Bull.}} \bibinfo{volume}{28}, \bibinfo{number}{2} (\bibinfo{year}{2005}), \bibinfo{pages}{11--16}.
\newblock
\urldef\tempurl%
\url{http://sites.computer.org/debull/A05june/stavros.ps}
\showURL{%
\tempurl}


\bibitem[Harizopoulos et~al\mbox{.}(2005)]%
        {qpipe}
\bibfield{author}{\bibinfo{person}{Stavros Harizopoulos}, \bibinfo{person}{Vladislav Shkapenyuk}, {and} \bibinfo{person}{Anastassia Ailamaki}.} \bibinfo{year}{2005}\natexlab{}.
\newblock \showarticletitle{QPipe: A Simultaneously Pipelined Relational Query Engine}. In \bibinfo{booktitle}{\emph{Proceedings of the 2005 ACM SIGMOD International Conference on Management of Data}} (Baltimore, Maryland) \emph{(\bibinfo{series}{SIGMOD '05})}. \bibinfo{publisher}{Association for Computing Machinery}, \bibinfo{address}{New York, NY, USA}, \bibinfo{pages}{383–394}.
\newblock
\showISBNx{1595930604}
\urldef\tempurl%
\url{https://doi.org/10.1145/1066157.1066201}
\showDOI{\tempurl}


\bibitem[iostat(2024)]%
        {IostatLinuxManual}
\bibfield{author}{\bibinfo{person}{iostat}.} \bibinfo{year}{2024}\natexlab{}.
\newblock
\newblock
\urldef\tempurl%
\url{https://linux.die.net/man/1/iostat}
\showURL{%
\tempurl}


\bibitem[Isenko et~al\mbox{.}(2022)]%
        {isenko}
\bibfield{author}{\bibinfo{person}{Alexander Isenko}, \bibinfo{person}{Ruben Mayer}, \bibinfo{person}{Jeffrey Jedele}, {and} \bibinfo{person}{Hans-Arno Jacobsen}.} \bibinfo{year}{2022}\natexlab{}.
\newblock \showarticletitle{Where Is My Training Bottleneck? Hidden Trade-Offs in Deep Learning Preprocessing Pipelines}. In \bibinfo{booktitle}{\emph{Proceedings of the 2022 International Conference on Management of Data}} (Philadelphia, PA, USA) \emph{(\bibinfo{series}{SIGMOD '22})}. \bibinfo{publisher}{Association for Computing Machinery}, \bibinfo{address}{New York, NY, USA}, \bibinfo{pages}{1825–1839}.
\newblock
\showISBNx{9781450392495}
\urldef\tempurl%
\url{https://doi.org/10.1145/3514221.3517848}
\showDOI{\tempurl}


\bibitem[Jeon et~al\mbox{.}(2019)]%
        {microsoftDLClusterStudy}
\bibfield{author}{\bibinfo{person}{Myeongjae Jeon}, \bibinfo{person}{Shivaram Venkataraman}, \bibinfo{person}{Amar Phanishayee}, \bibinfo{person}{Junjie Qian}, \bibinfo{person}{Wencong Xiao}, {and} \bibinfo{person}{Fan Yang}.} \bibinfo{year}{2019}\natexlab{}.
\newblock \showarticletitle{{Analysis of {Large-Scale} {Multi-Tenant} {GPU} Clusters for {DNN} Training Workloads}}. In \bibinfo{booktitle}{\emph{2019 USENIX Annual Technical Conference (USENIX ATC 19)}}. \bibinfo{publisher}{USENIX}, \bibinfo{address}{USA}, \bibinfo{pages}{947--960}.
\newblock


\bibitem[Johnson et~al\mbox{.}(2007)]%
        {JohnsonHPMHSAF07}
\bibfield{author}{\bibinfo{person}{Ryan Johnson}, \bibinfo{person}{Nikos Hardavellas}, \bibinfo{person}{Ippokratis Pandis}, \bibinfo{person}{Naju Mancheril}, \bibinfo{person}{Stavros Harizopoulos}, \bibinfo{person}{Kivanc Sabirli}, \bibinfo{person}{Anastassia Ailamaki}, {and} \bibinfo{person}{Babak Falsafi}.} \bibinfo{year}{2007}\natexlab{}.
\newblock \showarticletitle{To Share or Not To Share?}. In \bibinfo{booktitle}{\emph{Proceedings of the 33rd International Conference on Very Large Data Bases, University of Vienna, Austria, September 23-27, 2007}}, \bibfield{editor}{\bibinfo{person}{Christoph Koch}, \bibinfo{person}{Johannes Gehrke}, \bibinfo{person}{Minos~N. Garofalakis}, \bibinfo{person}{Divesh Srivastava}, \bibinfo{person}{Karl Aberer}, \bibinfo{person}{Anand Deshpande}, \bibinfo{person}{Daniela Florescu}, \bibinfo{person}{Chee~Yong Chan}, \bibinfo{person}{Venkatesh Ganti}, \bibinfo{person}{Carl{-}Christian Kanne}, \bibinfo{person}{Wolfgang Klas}, {and} \bibinfo{person}{Erich~J. Neuhold}} (Eds.). \bibinfo{publisher}{{ACM}}, \bibinfo{address}{Austria}, \bibinfo{pages}{351--362}.
\newblock
\urldef\tempurl%
\url{http://www.vldb.org/conf/2007/papers/research/p351-johnson.pdf}
\showURL{%
\tempurl}


\bibitem[Kakaraparthy et~al\mbox{.}(2019)]%
        {234827}
\bibfield{author}{\bibinfo{person}{Aarati Kakaraparthy}, \bibinfo{person}{Abhay Venkatesh}, \bibinfo{person}{Amar Phanishayee}, {and} \bibinfo{person}{Shivaram Venkataraman}.} \bibinfo{year}{2019}\natexlab{}.
\newblock \showarticletitle{The Case for Unifying Data Loading in Machine Learning Clusters}. In \bibinfo{booktitle}{\emph{11th USENIX Workshop on Hot Topics in Cloud Computing (HotCloud 19)}}. \bibinfo{publisher}{USENIX Association}, \bibinfo{address}{Renton, WA}, \bibinfo{pages}{12}.
\newblock
\urldef\tempurl%
\url{https://www.usenix.org/conference/hotcloud19/presentation/kakaraparthy}
\showURL{%
\tempurl}


\bibitem[kernel.org(2021)]%
        {topmanual}
kernel.org \bibinfo{year}{2021}\natexlab{}.
\newblock \bibinfo{booktitle}{\emph{{top(1) — Linux manual page}}}.
\newblock kernel.org.
\newblock
\newblock
\shownote{\url{https://man7.org/linux/man-pages/man1/top.1.html}}.


\bibitem[Kim et~al\mbox{.}(2024)]%
        {fusionflow}
\bibfield{author}{\bibinfo{person}{Taeyoon Kim}, \bibinfo{person}{ChanHo Park}, \bibinfo{person}{Heelim Hong}, \bibinfo{person}{Minseok Kim}, \bibinfo{person}{Ze Jin}, \bibinfo{person}{Changdae Kim}, \bibinfo{person}{Ji-Yong Shin}, {and} \bibinfo{person}{Myeongjae Jeo}.} \bibinfo{year}{2024}\natexlab{}.
\newblock \showarticletitle{{FusionFlow: Accelerating Data Preprocessing for Machine Learning with CPU-GPU Cooperation}}.
\newblock \bibinfo{journal}{\emph{Proc. VLDB Endow.}} \bibinfo{volume}{17}, \bibinfo{number}{4} (\bibinfo{year}{2024}), \bibinfo{pages}{863–876}.
\newblock
\showISSN{2150-8097}
\urldef\tempurl%
\url{https://doi.org/10.14778/3636218.3636238}
\showDOI{\tempurl}


\bibitem[Koliousis et~al\mbox{.}(2019)]%
        {crossbow}
\bibfield{author}{\bibinfo{person}{Alexandros Koliousis}, \bibinfo{person}{Pijika Watcharapichat}, \bibinfo{person}{Matthias Weidlich}, \bibinfo{person}{Luo Mai}, \bibinfo{person}{Paolo Costa}, {and} \bibinfo{person}{Peter Pietzuch}.} \bibinfo{year}{2019}\natexlab{}.
\newblock \showarticletitle{{Crossbow: Scaling Deep Learning with Small Batch Sizes on Multi-GPU Servers}}.
\newblock \bibinfo{journal}{\emph{PVLDB}} \bibinfo{volume}{12}, \bibinfo{number}{11} (\bibinfo{year}{2019}), \bibinfo{pages}{1399–1412}.
\newblock


\bibitem[Kuchnik et~al\mbox{.}(2022)]%
        {kuchnik_plumber_2022}
\bibfield{author}{\bibinfo{person}{Michael Kuchnik}, \bibinfo{person}{Ana Klimovic}, \bibinfo{person}{Jiri Simsa}, \bibinfo{person}{Virginia Smith}, {and} \bibinfo{person}{George Amvrosiadis}.} \bibinfo{year}{2022}\natexlab{}.
\newblock \showarticletitle{Plumber: {Diagnosing} and removing performance bottlenecks in machine learning data pipelines}.
\newblock \bibinfo{journal}{\emph{Proceedings of Machine Learning and Systems}}  \bibinfo{volume}{4} (\bibinfo{year}{2022}), \bibinfo{pages}{33--51}.
\newblock
\urldef\tempurl%
\url{https://proceedings.mlsys.org/paper_files/paper/2022/hash/d0e90e9a9310570dfa643aa3b2da6e89-Abstract.html}
\showURL{%
\tempurl}


\bibitem[Li et~al\mbox{.}(2020)]%
        {li2020massively}
\bibfield{author}{\bibinfo{person}{Liam Li}, \bibinfo{person}{Kevin Jamieson}, \bibinfo{person}{Afshin Rostamizadeh}, \bibinfo{person}{Ekaterina Gonina}, \bibinfo{person}{Jonathan Ben-tzur}, \bibinfo{person}{Moritz Hardt}, \bibinfo{person}{Benjamin Recht}, {and} \bibinfo{person}{Ameet Talwalkar}.} \bibinfo{year}{2020}\natexlab{}.
\newblock \showarticletitle{A System for Massively Parallel Hyperparameter Tuning}. In \bibinfo{booktitle}{\emph{Proceedings of Machine Learning and Systems}}, \bibfield{editor}{\bibinfo{person}{I.~Dhillon}, \bibinfo{person}{D.~Papailiopoulos}, {and} \bibinfo{person}{V.~Sze}} (Eds.), Vol.~\bibinfo{volume}{2}. \bibinfo{publisher}{MLSys}, \bibinfo{address}{USA}, \bibinfo{pages}{230--246}.
\newblock


\bibitem[Liu et~al\mbox{.}(2018)]%
        {liu2018progressive}
\bibfield{author}{\bibinfo{person}{Chenxi Liu}, \bibinfo{person}{Barret Zoph}, \bibinfo{person}{Maxim Neumann}, \bibinfo{person}{Jonathon Shlens}, \bibinfo{person}{Wei Hua}, \bibinfo{person}{Li-Jia Li}, \bibinfo{person}{Li Fei-Fei}, \bibinfo{person}{Alan Yuille}, \bibinfo{person}{Jonathan Huang}, {and} \bibinfo{person}{Kevin Murphy}.} \bibinfo{year}{2018}\natexlab{}.
\newblock \showarticletitle{Progressive neural architecture search}. In \bibinfo{booktitle}{\emph{Proceedings of the European conference on computer vision (ECCV)}}. \bibinfo{publisher}{Springer}, \bibinfo{address}{Germany}, \bibinfo{pages}{19--34}.
\newblock


\bibitem[LMPerf(2023)]%
        {mlcommonsstorage}
\bibfield{author}{\bibinfo{person}{LMPerf}.} \bibinfo{year}{2023}\natexlab{}.
\newblock \bibinfo{howpublished}{\url{https://mlcommons.org/benchmarks/storage/}}.
\newblock


\bibitem[Maschi and Alonso(2023)]%
        {10.1145/3592980.3595314}
\bibfield{author}{\bibinfo{person}{Fabio Maschi} {and} \bibinfo{person}{Gustavo Alonso}.} \bibinfo{year}{2023}\natexlab{}.
\newblock \showarticletitle{The Difficult Balance Between Modern Hardware and Conventional CPUs}. In \bibinfo{booktitle}{\emph{Proceedings of the 19th International Workshop on Data Management on New Hardware}} (Seattle, WA, USA) \emph{(\bibinfo{series}{DaMoN '23})}. \bibinfo{publisher}{Association for Computing Machinery}, \bibinfo{address}{New York, NY, USA}, \bibinfo{pages}{53–62}.
\newblock
\showISBNx{9798400701917}
\urldef\tempurl%
\url{https://doi.org/10.1145/3592980.3595314}
\showDOI{\tempurl}


\bibitem[{Microsoft Azure}(2023)]%
        {azure_ratio}
\bibfield{author}{\bibinfo{person}{{Microsoft Azure}}.} \bibinfo{year}{2023}\natexlab{}.
\newblock \bibinfo{howpublished}{\url{https://azure.microsoft.com/en-us/products/virtual-machines}}.
\newblock
\newblock
\shownote{Accessed: 2023-11-30}.


\bibitem[{Microsoft Research}(2024)]%
        {coordlcodebase}
\bibfield{author}{\bibinfo{person}{{Microsoft Research}}.} \bibinfo{year}{2024}\natexlab{}.
\newblock \bibinfo{howpublished}{\url{https://github.com/msr-fiddle/CoorDL/tree/master}}.
\newblock
\newblock
\shownote{Accessed: 2024-07-10}.


\bibitem[Mohan et~al\mbox{.}(2021)]%
        {mohan2021analyzing}
\bibfield{author}{\bibinfo{person}{Jayashree Mohan}, \bibinfo{person}{Amar Phanishayee}, \bibinfo{person}{Ashish Raniwala}, {and} \bibinfo{person}{Vijay Chidambaram}.} \bibinfo{year}{2021}\natexlab{}.
\newblock \showarticletitle{Analyzing and mitigating data stalls in DNN training}.
\newblock \bibinfo{journal}{\emph{Proc. VLDB Endow.}} \bibinfo{volume}{14}, \bibinfo{number}{5} (\bibinfo{date}{jan} \bibinfo{year}{2021}), \bibinfo{pages}{771–784}.
\newblock
\showISSN{2150-8097}
\urldef\tempurl%
\url{https://doi.org/10.14778/3446095.3446100}
\showDOI{\tempurl}


\bibitem[Murray et~al\mbox{.}(2021)]%
        {tf.data}
\bibfield{author}{\bibinfo{person}{Derek~G. Murray}, \bibinfo{person}{Ji\v{r}\'{\i} \v{S}im\v{s}a}, \bibinfo{person}{Ana Klimovic}, {and} \bibinfo{person}{Ihor Indyk}.} \bibinfo{year}{2021}\natexlab{}.
\newblock \showarticletitle{Tf.Data: A Machine Learning Data Processing Framework}.
\newblock \bibinfo{journal}{\emph{Proc. VLDB Endow.}} \bibinfo{volume}{14}, \bibinfo{number}{12} (\bibinfo{date}{jul} \bibinfo{year}{2021}), \bibinfo{pages}{2945–2958}.
\newblock
\showISSN{2150-8097}
\urldef\tempurl%
\url{https://doi.org/10.14778/3476311.3476374}
\showDOI{\tempurl}


\bibitem[Nakandala et~al\mbox{.}(2020)]%
        {cerebro}
\bibfield{author}{\bibinfo{person}{Supun Nakandala}, \bibinfo{person}{Yuhao Zhang}, {and} \bibinfo{person}{Arun Kumar}.} \bibinfo{year}{2020}\natexlab{}.
\newblock \showarticletitle{Cerebro: A Data System for Optimized Deep Learning Model Selection}.
\newblock \bibinfo{journal}{\emph{Proc. VLDB Endow.}} \bibinfo{volume}{13}, \bibinfo{number}{12} (\bibinfo{date}{jul} \bibinfo{year}{2020}), \bibinfo{pages}{2159–2173}.
\newblock
\showISSN{2150-8097}
\urldef\tempurl%
\url{https://doi.org/10.14778/3407790.3407816}
\showDOI{\tempurl}


\bibitem[Nikas et~al\mbox{.}(2019)]%
        {dicer}
\bibfield{author}{\bibinfo{person}{Konstantinos Nikas}, \bibinfo{person}{Nikela Papadopoulou}, \bibinfo{person}{Dimitra Giantsidi}, \bibinfo{person}{Vasileios Karakostas}, \bibinfo{person}{Georgios Goumas}, {and} \bibinfo{person}{Nectarios Koziris}.} \bibinfo{year}{2019}\natexlab{}.
\newblock \showarticletitle{{DICER: Diligent Cache Partitioning for Efficient Workload Consolidation}}. In \bibinfo{booktitle}{\emph{ICPP}}. \bibinfo{publisher}{Association for Computing Machinery}, \bibinfo{address}{Japan}, \bibinfo{pages}{15:1--15:10}.
\newblock


\bibitem[{NVIDIA}(2012)]%
        {NVIDIASystemManagement2012}
\bibfield{author}{\bibinfo{person}{{NVIDIA}}.} \bibinfo{year}{2012}\natexlab{}.
\newblock
\newblock
\urldef\tempurl%
\url{https://developer.nvidia.com/nvidia-system-management-interface}
\showURL{%
\tempurl}


\bibitem[NVIDIA(2022)]%
        {dcgmmanual}
\bibfield{author}{\bibinfo{person}{NVIDIA}.} \bibinfo{year}{2022}\natexlab{}.
\newblock \bibinfo{booktitle}{\emph{{Data Center {GPU} Manager Documentation}}}.
\newblock \bibinfo{type}{{T}echnical {R}eport}. \bibinfo{institution}{NVIDIA}.
\newblock
\newblock
\shownote{\url{https://docs.nvidia.com/datacenter/dcgm/latest/dcgm-user-guide/}}.


\bibitem[NVIDIA(2023)]%
        {nvidia_dali}
\bibfield{author}{\bibinfo{person}{NVIDIA}.} \bibinfo{year}{2023}\natexlab{}.
\newblock \bibinfo{howpublished}{\url{https://github.com/NVIDIA/DALI}}.
\newblock
\newblock
\shownote{Accessed: 2023-05-19}.


\bibitem[{NVIDIA}(2023)]%
        {nvidia_mps_doc}
\bibfield{author}{\bibinfo{person}{{NVIDIA}}.} \bibinfo{year}{2023}\natexlab{}.
\newblock \bibinfo{howpublished}{\url{https://docs.nvidia.com/deploy/mps/index.html}}.
\newblock
\newblock
\shownote{Accessed: 2023-05-27}.


\bibitem[Panayotov et~al\mbox{.}(2015)]%
        {7178964}
\bibfield{author}{\bibinfo{person}{Vassil Panayotov}, \bibinfo{person}{Guoguo Chen}, \bibinfo{person}{Daniel Povey}, {and} \bibinfo{person}{Sanjeev Khudanpur}.} \bibinfo{year}{2015}\natexlab{}.
\newblock \showarticletitle{Librispeech: An ASR corpus based on public domain audio books}. In \bibinfo{booktitle}{\emph{2015 IEEE International Conference on Acoustics, Speech and Signal Processing (ICASSP)}}. \bibinfo{publisher}{IEE}, \bibinfo{address}{USA}, \bibinfo{pages}{5206--5210}.
\newblock
\urldef\tempurl%
\url{https://doi.org/10.1109/ICASSP.2015.7178964}
\showDOI{\tempurl}


\bibitem[Paszke et~al\mbox{.}(2019)]%
        {pytorch}
\bibfield{author}{\bibinfo{person}{Adam Paszke}, \bibinfo{person}{Sam Gross}, \bibinfo{person}{Francisco Massa}, \bibinfo{person}{Adam Lerer}, \bibinfo{person}{James Bradbury}, \bibinfo{person}{Gregory Chanan}, \bibinfo{person}{Trevor Killeen}, \bibinfo{person}{Zeming Lin}, \bibinfo{person}{Natalia Gimelshein}, \bibinfo{person}{Luca Antiga}, \bibinfo{person}{Alban Desmaison}, \bibinfo{person}{Andreas Kopf}, \bibinfo{person}{Edward Yang}, \bibinfo{person}{Zachary DeVito}, \bibinfo{person}{Martin Raison}, \bibinfo{person}{Alykhan Tejani}, \bibinfo{person}{Sasank Chilamkurthy}, \bibinfo{person}{Benoit Steiner}, \bibinfo{person}{Lu Fang}, \bibinfo{person}{Junjie Bai}, {and} \bibinfo{person}{Soumith Chintala}.} \bibinfo{year}{2019}\natexlab{}.
\newblock \showarticletitle{PyTorch: An Imperative Style, High-Performance Deep Learning Library}.
\newblock In \bibinfo{booktitle}{\emph{Advances in Neural Information Processing Systems 32}}. \bibinfo{publisher}{Curran Associates, Inc.}, \bibinfo{address}{Canada}, \bibinfo{pages}{8024--8035}.
\newblock
\urldef\tempurl%
\url{http://papers.neurips.cc/paper/9015-pytorch-an-imperative-style-high-performance-deep-learning-library.pdf}
\showURL{%
\tempurl}


\bibitem[Patterson et~al\mbox{.}(2022)]%
        {googleCarbon}
\bibfield{author}{\bibinfo{person}{David~A. Patterson}, \bibinfo{person}{Joseph Gonzalez}, \bibinfo{person}{Urs H{\"{o}}lzle}, \bibinfo{person}{Quoc~V. Le}, \bibinfo{person}{Chen Liang}, \bibinfo{person}{Lluis{-}Miquel Munguia}, \bibinfo{person}{Daniel Rothchild}, \bibinfo{person}{David~R. So}, \bibinfo{person}{Maud Texier}, {and} \bibinfo{person}{Jeff Dean}.} \bibinfo{year}{2022}\natexlab{}.
\newblock \showarticletitle{The Carbon Footprint of Machine Learning Training Will Plateau, Then Shrink}.
\newblock \bibinfo{journal}{\emph{Computer}} \bibinfo{volume}{55}, \bibinfo{number}{7} (\bibinfo{year}{2022}), \bibinfo{pages}{18--28}.
\newblock
\urldef\tempurl%
\url{https://doi.org/10.1109/MC.2022.3148714}
\showDOI{\tempurl}


\bibitem[Probst et~al\mbox{.}(2019)]%
        {probst2018tunability}
\bibfield{author}{\bibinfo{person}{Philipp Probst}, \bibinfo{person}{Anne-Laure Boulesteix}, {and} \bibinfo{person}{Bernd Bischl}.} \bibinfo{year}{2019}\natexlab{}.
\newblock \showarticletitle{{Tunability: Importance of hyperparameters of machine learning algorithms}}.
\newblock \bibinfo{journal}{\emph{Journal of Machine Learning Research}} \bibinfo{volume}{20}, \bibinfo{number}{1} (\bibinfo{date}{jan} \bibinfo{year}{2019}), \bibinfo{pages}{1934--1965}.
\newblock
\showISSN{1532-4435}


\bibitem[Psaroudakis et~al\mbox{.}(2013)]%
        {psaroudakisAA13}
\bibfield{author}{\bibinfo{person}{Iraklis Psaroudakis}, \bibinfo{person}{Manos Athanassoulis}, {and} \bibinfo{person}{Anastasia Ailamaki}.} \bibinfo{year}{2013}\natexlab{}.
\newblock \showarticletitle{Sharing data and work across concurrent analytical queries}.
\newblock \bibinfo{journal}{\emph{Proc. VLDB Endow.}} \bibinfo{volume}{6}, \bibinfo{number}{9} (\bibinfo{date}{jul} \bibinfo{year}{2013}), \bibinfo{pages}{637–648}.
\newblock
\showISSN{2150-8097}
\urldef\tempurl%
\url{https://doi.org/10.14778/2536360.2536364}
\showDOI{\tempurl}


\bibitem[Ramesh et~al\mbox{.}(2022)]%
        {ramesh2022hierarchical}
\bibfield{author}{\bibinfo{person}{Aditya Ramesh}, \bibinfo{person}{Prafulla Dhariwal}, \bibinfo{person}{Alex Nichol}, \bibinfo{person}{Casey Chu}, {and} \bibinfo{person}{Mark Chen}.} \bibinfo{year}{2022}\natexlab{}.
\newblock
\newblock
\showeprint[arxiv]{2204.06125}~[cs.CV]


\bibitem[Robroek et~al\mbox{.}(2023)]%
        {radt}
\bibfield{author}{\bibinfo{person}{Ties Robroek}, \bibinfo{person}{Aaron Duane}, \bibinfo{person}{Ehsan Yousefzadeh-Asl-Miandoab}, {and} \bibinfo{person}{Pinar Tozun}.} \bibinfo{year}{2023}\natexlab{}.
\newblock \showarticletitle{Data Management and Visualization for Benchmarking Deep Learning Training Systems}. In \bibinfo{booktitle}{\emph{Proceedings of the Seventh Workshop on Data Management for End-to-End Machine Learning}}. \bibinfo{publisher}{Association for Computing Machinery}, \bibinfo{address}{USA}, \bibinfo{pages}{1--5}.
\newblock


\bibitem[Robroek et~al\mbox{.}(2024)]%
        {robroek2023analysis}
\bibfield{author}{\bibinfo{person}{Ties Robroek}, \bibinfo{person}{Ehsan Yousefzadeh{-}Asl{-}Miandoab}, {and} \bibinfo{person}{Pinar T{\"{o}}z{\"{u}}n}.} \bibinfo{year}{2024}\natexlab{}.
\newblock \showarticletitle{An Analysis of Collocation on GPUs for Deep Learning Training}. In \bibinfo{booktitle}{\emph{Proceedings of the 4th Workshop on Machine Learning and Systems, EuroMLSys 2024, Athens, Greece, 22 April 2024}}. \bibinfo{publisher}{{ACM}}, \bibinfo{address}{Greece}, \bibinfo{pages}{81--90}.
\newblock
\urldef\tempurl%
\url{https://doi.org/10.1145/3642970.3655827}
\showDOI{\tempurl}


\bibitem[Schwartz et~al\mbox{.}(2020)]%
        {greenai}
\bibfield{author}{\bibinfo{person}{Roy Schwartz}, \bibinfo{person}{Jesse Dodge}, \bibinfo{person}{Noah~A. Smith}, {and} \bibinfo{person}{Oren Etzioni}.} \bibinfo{year}{2020}\natexlab{}.
\newblock \showarticletitle{Green AI}.
\newblock \bibinfo{journal}{\emph{Commun. ACM}} \bibinfo{volume}{63}, \bibinfo{number}{12} (\bibinfo{date}{nov} \bibinfo{year}{2020}), \bibinfo{pages}{54–63}.
\newblock
\showISSN{0001-0782}
\urldef\tempurl%
\url{https://doi.org/10.1145/3381831}
\showDOI{\tempurl}


\bibitem[Sharma et~al\mbox{.}(2018)]%
        {sharma-etal-2018-conceptual}
\bibfield{author}{\bibinfo{person}{Piyush Sharma}, \bibinfo{person}{Nan Ding}, \bibinfo{person}{Sebastian Goodman}, {and} \bibinfo{person}{Radu Soricut}.} \bibinfo{year}{2018}\natexlab{}.
\newblock \showarticletitle{Conceptual Captions: A Cleaned, Hypernymed, Image Alt-text Dataset For Automatic Image Captioning}. In \bibinfo{booktitle}{\emph{Proceedings of the 56th Annual Meeting of the Association for Computational Linguistics (Volume 1: Long Papers)}}, \bibfield{editor}{\bibinfo{person}{Iryna Gurevych} {and} \bibinfo{person}{Yusuke Miyao}} (Eds.). \bibinfo{publisher}{Association for Computational Linguistics}, \bibinfo{address}{Melbourne, Australia}, \bibinfo{pages}{2556--2565}.
\newblock
\urldef\tempurl%
\url{https://doi.org/10.18653/v1/P18-1238}
\showDOI{\tempurl}


\bibitem[Spijkervet and Burgoyne(2021)]%
        {spijkervet2021contrastive}
\bibfield{author}{\bibinfo{person}{Janne Spijkervet} {and} \bibinfo{person}{John~Ashley Burgoyne}.} \bibinfo{year}{2021}\natexlab{}.
\newblock , \bibinfo{numpages}{673-681}~pages.
\newblock
\urldef\tempurl%
\url{https://doi.org/10.5281/zenodo.5624573}
\showDOI{\tempurl}


\bibitem[Strati et~al\mbox{.}(2024)]%
        {orion}
\bibfield{author}{\bibinfo{person}{Foteini Strati}, \bibinfo{person}{Xianzhe Ma}, {and} \bibinfo{person}{Ana Klimovic}.} \bibinfo{year}{2024}\natexlab{}.
\newblock \showarticletitle{Orion: Interference-aware, Fine-grained {GPU} Sharing for {ML} Applications}. In \bibinfo{booktitle}{\emph{Proceedings of the Nineteenth European Conference on Computer Systems, EuroSys 2024, Athens, Greece, April 22-25, 2024}}. \bibinfo{publisher}{{ACM}}, \bibinfo{address}{Greece}, \bibinfo{pages}{1075--1092}.
\newblock
\urldef\tempurl%
\url{https://doi.org/10.1145/3627703.3629578}
\showDOI{\tempurl}


\bibitem[Strubell et~al\mbox{.}(2019)]%
        {StrubellGM19}
\bibfield{author}{\bibinfo{person}{Emma Strubell}, \bibinfo{person}{Ananya Ganesh}, {and} \bibinfo{person}{Andrew McCallum}.} \bibinfo{year}{2019}\natexlab{}.
\newblock \showarticletitle{Energy and Policy Considerations for Deep Learning in {NLP}}. In \bibinfo{booktitle}{\emph{Proceedings of the 57th Conference of the Association for Computational Linguistics, {ACL} 2019, Florence, Italy, July 28- August 2, 2019, Volume 1: Long Papers}}, \bibfield{editor}{\bibinfo{person}{Anna Korhonen}, \bibinfo{person}{David~R. Traum}, {and} \bibinfo{person}{Llu{\'{\i}}s M{\`{a}}rquez}} (Eds.). \bibinfo{publisher}{Association for Computational Linguistics}, \bibinfo{address}{Italy}, \bibinfo{pages}{3645--3650}.
\newblock
\urldef\tempurl%
\url{https://doi.org/10.18653/V1/P19-1355}
\showDOI{\tempurl}


\bibitem[Taori et~al\mbox{.}(2023)]%
        {alpaca}
\bibfield{author}{\bibinfo{person}{Rohan Taori}, \bibinfo{person}{Ishaan Gulrajani}, \bibinfo{person}{Tianyi Zhang}, \bibinfo{person}{Yann Dubois}, \bibinfo{person}{Xuechen Li}, \bibinfo{person}{Carlos Guestrin}, \bibinfo{person}{Percy Liang}, {and} \bibinfo{person}{Tatsunori~B. Hashimoto}.} \bibinfo{year}{2023}\natexlab{}.
\newblock \bibinfo{howpublished}{\url{https://github.com/tatsu-lab/stanford_alpaca}}.
\newblock


\bibitem[torchtune maintainers and contributors(2024)]%
        {torchtune}
\bibfield{author}{\bibinfo{person}{torchtune maintainers} {and} \bibinfo{person}{contributors}.} \bibinfo{year}{2024}\natexlab{}.
\newblock \bibinfo{booktitle}{\emph{torchtune: PyTorch's finetuning library}}.
\newblock PyTorch.
\newblock
\urldef\tempurl%
\url{https//github.com/pytorch/torchtune}
\showURL{%
\tempurl}


\bibitem[T\"{o}z\"{u}n et~al\mbox{.}(2014)]%
        {addict}
\bibfield{author}{\bibinfo{person}{Pinar T\"{o}z\"{u}n}, \bibinfo{person}{Islam Atta}, \bibinfo{person}{Anastasia Ailamaki}, {and} \bibinfo{person}{Andreas Moshovos}.} \bibinfo{year}{2014}\natexlab{}.
\newblock \showarticletitle{ADDICT: advanced instruction chasing for transactions}.
\newblock \bibinfo{journal}{\emph{Proc. VLDB Endow.}} \bibinfo{volume}{7}, \bibinfo{number}{14} (\bibinfo{date}{oct} \bibinfo{year}{2014}), \bibinfo{pages}{1893–1904}.
\newblock
\showISSN{2150-8097}
\urldef\tempurl%
\url{https://doi.org/10.14778/2733085.2733095}
\showDOI{\tempurl}


\bibitem[Um et~al\mbox{.}(2023)]%
        {fastflow}
\bibfield{author}{\bibinfo{person}{Taegeon Um}, \bibinfo{person}{Byungsoo Oh}, \bibinfo{person}{Byeongchan Seo}, \bibinfo{person}{Minhyeok Kweun}, \bibinfo{person}{Goeun Kim}, {and} \bibinfo{person}{Woo-Yeon Lee}.} \bibinfo{year}{2023}\natexlab{}.
\newblock \showarticletitle{{FastFlow: Accelerating Deep Learning Model Training with Smart Offloading of Input Data Pipeline}}.
\newblock \bibinfo{journal}{\emph{Proc. VLDB Endow.}} \bibinfo{volume}{16}, \bibinfo{number}{5} (\bibinfo{date}{jan} \bibinfo{year}{2023}), \bibinfo{pages}{1086–1099}.
\newblock
\showISSN{2150-8097}
\urldef\tempurl%
\url{https://doi.org/10.14778/3579075.3579083}
\showDOI{\tempurl}


\bibitem[Varoquaux et~al\mbox{.}(2025)]%
        {hypesustainability}
\bibfield{author}{\bibinfo{person}{Ga{\"{e}}l Varoquaux}, \bibinfo{person}{Sasha Luccioni}, {and} \bibinfo{person}{Meredith Whittaker}.} \bibinfo{year}{2025}\natexlab{}.
\newblock \showarticletitle{Hype, Sustainability, and the Price of the Bigger-is-Better Paradigm in {AI}}. In \bibinfo{booktitle}{\emph{Proceedings of the 2025 {ACM} Conference on Fairness, Accountability, and Transparency, FAccT 2025, Athens, Greece, June 23-26, 2025}}. \bibinfo{publisher}{{ACM}}, \bibinfo{address}{Athens, Canada}, \bibinfo{pages}{61--75}.
\newblock
\urldef\tempurl%
\url{https://doi.org/10.1145/3715275.3732006}
\showDOI{\tempurl}


\bibitem[Ventura et~al\mbox{.}(2021)]%
        {VenturaKQM21}
\bibfield{author}{\bibinfo{person}{Francesco Ventura}, \bibinfo{person}{Zoi Kaoudi}, \bibinfo{person}{Jorge{-}Arnulfo Quian{\'{e}}{-}Ruiz}, {and} \bibinfo{person}{Volker Markl}.} \bibinfo{year}{2021}\natexlab{}.
\newblock \showarticletitle{Expand your Training Limits! Generating Training Data for ML-based Data Management}. In \bibinfo{booktitle}{\emph{{SIGMOD} '21: International Conference on Management of Data, Virtual Event, China, June 20-25, 2021}}, \bibfield{editor}{\bibinfo{person}{Guoliang Li}, \bibinfo{person}{Zhanhuai Li}, \bibinfo{person}{Stratos Idreos}, {and} \bibinfo{person}{Divesh Srivastava}} (Eds.). \bibinfo{publisher}{{ACM}}, \bibinfo{address}{China}, \bibinfo{pages}{1865--1878}.
\newblock
\urldef\tempurl%
\url{https://doi.org/10.1145/3448016.3457286}
\showDOI{\tempurl}


\bibitem[Wang(2022)]%
        {pw2022dalle2}
\bibfield{author}{\bibinfo{person}{Phil Wang}.} \bibinfo{year}{2022}\natexlab{}.
\newblock \bibinfo{howpublished}{\url{https://github.com/lucidrains/DALLE2-pytorch}}.
\newblock


\bibitem[Wang et~al\mbox{.}(2021)]%
        {hfta}
\bibfield{author}{\bibinfo{person}{Shang Wang}, \bibinfo{person}{Peiming Yang}, \bibinfo{person}{Yuxuan Zheng}, \bibinfo{person}{Xin Li}, {and} \bibinfo{person}{Gennady Pekhimenko}.} \bibinfo{year}{2021}\natexlab{}.
\newblock \showarticletitle{{Horizontally Fused Training Array: An Effective Hardware Utilization Squeezer for Training Novel Deep Learning Models}}.
\newblock \bibinfo{journal}{\emph{Proceedings of Machine Learning and Systems}}  \bibinfo{volume}{3} (\bibinfo{year}{2021}), \bibinfo{pages}{599--623}.
\newblock


\bibitem[Weng et~al\mbox{.}(2022)]%
        {alibabaStudyGPU}
\bibfield{author}{\bibinfo{person}{Qizhen Weng}, \bibinfo{person}{Wencong Xiao}, \bibinfo{person}{Yinghao Yu}, \bibinfo{person}{Wei Wang}, \bibinfo{person}{Cheng Wang}, \bibinfo{person}{Jian He}, \bibinfo{person}{Yong Li}, \bibinfo{person}{Liping Zhang}, \bibinfo{person}{Wei Lin}, {and} \bibinfo{person}{Yu Ding}.} \bibinfo{year}{2022}\natexlab{}.
\newblock \showarticletitle{{{MLaaS} in the Wild: Workload Analysis and Scheduling in {Large-Scale} Heterogeneous {GPU} Clusters}}. In \bibinfo{booktitle}{\emph{19th USENIX Symposium on Networked Systems Design and Implementation (NSDI 22)}}. \bibinfo{publisher}{USENIX Association}, \bibinfo{address}{USA}, \bibinfo{pages}{945--960}.
\newblock
\urldef\tempurl%
\url{https://www.usenix.org/conference/nsdi22/presentation/weng}
\showURL{%
\tempurl}


\bibitem[Wightman(2019)]%
        {rw2019timm}
\bibfield{author}{\bibinfo{person}{Ross Wightman}.} \bibinfo{year}{2019}\natexlab{}.
\newblock \bibinfo{howpublished}{\url{https://github.com/rwightman/pytorch-image-models}}.
\newblock
\urldef\tempurl%
\url{https://doi.org/10.5281/zenodo.4414861}
\showDOI{\tempurl}


\bibitem[Xu et~al\mbox{.}(2022)]%
        {xuDeepLearningDataloader2022}
\bibfield{author}{\bibinfo{person}{Jingwei Xu}, \bibinfo{person}{Guochang Wang}, \bibinfo{person}{Yuan Yao}, \bibinfo{person}{Zenan Li}, \bibinfo{person}{Chun Cao}, {and} \bibinfo{person}{Hanghang Tong}.} \bibinfo{year}{2022}\natexlab{}.
\newblock \showarticletitle{A {{Deep Learning Dataloader}} with {{Shared Data Preparation}}}.
\newblock \bibinfo{journal}{\emph{Advances in Neural Information Processing Systems}}  \bibinfo{volume}{35} (\bibinfo{year}{2022}), \bibinfo{pages}{17146--17156}.
\newblock


\bibitem[Xu et~al\mbox{.}(2024)]%
        {joadercodebase}
\bibfield{author}{\bibinfo{person}{Jingwei Xu}, \bibinfo{person}{Guochang Wang}, \bibinfo{person}{Yuan Yao}, \bibinfo{person}{Zenan Li}, \bibinfo{person}{Chun Cao}, {and} \bibinfo{person}{Hanghang Tong}.} \bibinfo{year}{2024}\natexlab{}.
\newblock \bibinfo{howpublished}{\url{https://github.com/XieJiann/Joader}}.
\newblock
\newblock
\shownote{Accessed: 2024-07-10}.


\bibitem[Yang et~al\mbox{.}(2025)]%
        {yang2024qwen25}
\bibfield{author}{\bibinfo{person}{Qwen:~An Yang}, \bibinfo{person}{Baosong Yang}, \bibinfo{person}{Beichen Zhang}, \bibinfo{person}{Binyuan Hui}, \bibinfo{person}{Bo Zheng}, \bibinfo{person}{Bowen Yu}, \bibinfo{person}{Chengyuan Li}, \bibinfo{person}{Dayiheng Liu}, \bibinfo{person}{Fei Huang}, \bibinfo{person}{Haoran Wei}, \bibinfo{person}{Huan Lin}, \bibinfo{person}{Jian Yang}, \bibinfo{person}{Jianhong Tu}, \bibinfo{person}{Jianwei Zhang}, \bibinfo{person}{Jianxin Yang}, \bibinfo{person}{Jiaxi Yang}, \bibinfo{person}{Jingren Zhou}, \bibinfo{person}{Junyang Lin}, \bibinfo{person}{Kai Dang}, \bibinfo{person}{Keming Lu}, \bibinfo{person}{Keqin Bao}, \bibinfo{person}{Kexin Yang}, \bibinfo{person}{Le Yu}, \bibinfo{person}{Mei Li}, \bibinfo{person}{Mingfeng Xue}, \bibinfo{person}{Pei Zhang}, \bibinfo{person}{Qin Zhu}, \bibinfo{person}{Rui Men}, \bibinfo{person}{Runji Lin}, \bibinfo{person}{Tianhao Li}, \bibinfo{person}{Tianyi Tang}, \bibinfo{person}{Tingyu Xia}, \bibinfo{person}{Xingzhang Ren},
  \bibinfo{person}{Xuancheng Ren}, \bibinfo{person}{Yang Fan}, \bibinfo{person}{Yang Su}, \bibinfo{person}{Yichang Zhang}, \bibinfo{person}{Yu Wan}, \bibinfo{person}{Yuqiong Liu}, \bibinfo{person}{Zeyu Cui}, \bibinfo{person}{Zhenru Zhang}, {and} \bibinfo{person}{Zihan Qiu}.} \bibinfo{year}{2025}\natexlab{}.
\newblock \bibinfo{title}{Qwen2.5 Technical Report}.
\newblock
\newblock
\showeprint[arxiv]{2412.15115}~[cs.CL]
\urldef\tempurl%
\url{https://arxiv.org/abs/2412.15115}
\showURL{%
\tempurl}


\bibitem[Yousefzadeh{-}Asl{-}Miandoab et~al\mbox{.}(2023)]%
        {yousefzadeh2023profiling}
\bibfield{author}{\bibinfo{person}{Ehsan Yousefzadeh{-}Asl{-}Miandoab}, \bibinfo{person}{Ties Robroek}, {and} \bibinfo{person}{Pinar T{\"{o}}z{\"{u}}n}.} \bibinfo{year}{2023}\natexlab{}.
\newblock \showarticletitle{Profiling and Monitoring Deep Learning Training Tasks}. In \bibinfo{booktitle}{\emph{Proceedings of the 3rd Workshop on Machine Learning and Systems, EuroMLSys 2023, Rome, Italy, 8 May 2023}}, \bibfield{editor}{\bibinfo{person}{Eiko Yoneki} {and} \bibinfo{person}{Luigi Nardi}} (Eds.). \bibinfo{publisher}{{ACM}}, \bibinfo{address}{Italy}, \bibinfo{pages}{18--25}.
\newblock
\urldef\tempurl%
\url{https://doi.org/10.1145/3578356.3592589}
\showDOI{\tempurl}


\bibitem[{ZeroMQ}(2023)]%
        {zeromq}
\bibfield{author}{\bibinfo{person}{{ZeroMQ}}.} \bibinfo{year}{2023}\natexlab{}.
\newblock \bibinfo{howpublished}{\url{https://zeromq.org/}}.
\newblock
\newblock
\shownote{Accessed: 2023-12-11}.


\bibitem[Zhao et~al\mbox{.}(2024)]%
        {zhao_cedar_2024}
\bibfield{author}{\bibinfo{person}{Mark Zhao}, \bibinfo{person}{Emanuel Adamiak}, {and} \bibinfo{person}{Christos Kozyrakis}.} \bibinfo{year}{2024}\natexlab{}.
\newblock
\newblock
\urldef\tempurl%
\url{https://doi.org/10.48550/arXiv.2401.08895}
\showDOI{\tempurl}
\newblock
\shownote{arXiv:2401.08895 [cs]}.


\bibitem[Zhao et~al\mbox{.}(2022)]%
        {metaDataLoadingAnalysis}
\bibfield{author}{\bibinfo{person}{Mark Zhao}, \bibinfo{person}{Niket Agarwal}, \bibinfo{person}{Aarti Basant}, \bibinfo{person}{Bu\u{g}ra Gedik}, \bibinfo{person}{Satadru Pan}, \bibinfo{person}{Mustafa Ozdal}, \bibinfo{person}{Rakesh Komuravelli}, \bibinfo{person}{Jerry Pan}, \bibinfo{person}{Tianshu Bao}, \bibinfo{person}{Haowei Lu}, \bibinfo{person}{Sundaram Narayanan}, \bibinfo{person}{Jack Langman}, \bibinfo{person}{Kevin Wilfong}, \bibinfo{person}{Harsha Rastogi}, \bibinfo{person}{Carole-Jean Wu}, \bibinfo{person}{Christos Kozyrakis}, {and} \bibinfo{person}{Parik Pol}.} \bibinfo{year}{2022}\natexlab{}.
\newblock \showarticletitle{Understanding data storage and ingestion for large-scale deep recommendation model training: industrial product}. In \bibinfo{booktitle}{\emph{Proceedings of the 49th Annual International Symposium on Computer Architecture}} (New York, New York) \emph{(\bibinfo{series}{ISCA '22})}. \bibinfo{publisher}{Association for Computing Machinery}, \bibinfo{address}{New York, NY, USA}, \bibinfo{pages}{1042–1057}.
\newblock
\showISBNx{9781450386104}
\urldef\tempurl%
\url{https://doi.org/10.1145/3470496.3533044}
\showDOI{\tempurl}


\end{thebibliography}
